%% file: PecMan.tex
\documentclass[times, review, 10pt]{elsarticle}

\usepackage{newtxtext,newtxmath}
\usepackage{setspace}
\usepackage[
top=4.3cm,
bottom=4.3cm,
left=4.8cm,
right=4.8cm
]{geometry}

\usepackage{caption}
\renewcommand{\footnotesize}{\fontsize{8}{9}\selectfont}

\usepackage[dvipsnames, table]{xcolor}
\usepackage{graphicx, verbatim}
\usepackage{amsmath}
\usepackage{amsfonts}
\usepackage[normalem]{ulem}
\usepackage{booktabs}
\usepackage{multirow}
\usepackage[caption=false]{subfig}
\usepackage{color, colortbl, xcolor}
\usepackage{pifont}
\usepackage{tikz}
\usetikzlibrary{
    positioning,
    shapes.geometric,
    arrows.meta
}
\usepackage{pgfplots}
\pgfplotsset{compat=1.18}
\usetikzlibrary{external}
\AddToHook{env/algorithmic/begin}{\tikzexternaldisable}

\usepackage[pagebackref=false, breaklinks, colorlinks, citecolor=ForestGreen, linkcolor=BrickRed]{hyperref}
\usepackage[capitalize]{cleveref}

\usepackage{dirtytalk}
\usepackage{pgfplots}
\pgfplotsset{compat=newest}

\usepackage[dvipsnames]{xcolor}
\usepgfplotslibrary{fillbetween}
\usepackage{pgfplotstable}
\usepackage{hyperref}

\journal{Pattern Recognition}

\begin{document}
\title{People-Centred Medical Image Analysis via Fairness-Aware Human–AI Cooperation}

\begin{frontmatter}

%% Title, authors and addresses

%% use the tnoteref command within \title for footnotes;
%% use the tnotetext command for theassociated footnote;
%% use the fnref command within \author or \affiliation for footnotes;
%% use the fntext command for theassociated footnote;
%% use the corref command within \author for corresponding author footnotes;
%% use the cortext command for theassociated footnote;
%% use the ead command for the email address,
%% and the form \ead[url] for the home page:
%% \title{Title\tnoteref{label1}}
%% \tnotetext[label1]{}
%% \author{Name\corref{cor1}\fnref{label2}}
%% \ead{email address}
%% \ead[url]{home page}
%% \fntext[label2]{}
%% \cortext[cor1]{}
%% \affiliation{organization={},
%%             addressline={},
%%             city={},
%%             postcode={},
%%             state={},
%%             country={}}
%% \fntext[label3]{}

\title{}

%% use optional labels to link authors explicitly to addresses:
%% \author[label1,label2]{}
%% \affiliation[label1]{organization={},
%%             addressline={},
%%             city={},
%%             postcode={},
%%             state={},
%%             country={}}
%%
%% \affiliation[label2]{organization={},
%%             addressline={},
%%             city={},
%%             postcode={},
%%             state={},
%%             country={}}

\author[label1]{Zheng Zhang} %% Author name
\author[label1]{Milad Masroor} %% Author name
\author[label1]{Cuong Nguyen}
\author[label3]{Tahir Hassan}
\author[label5]{Yuanhong Chen}
\author[label2]{David Rosewarne}
\author[label1]{Kevin Wells}
\author[label4]{Thanh-Toan Do}
\author[label1]{Gustavo~Carneiro}

%% Author affiliation
\affiliation[label1]{organization={CVSSP, PAI, University of Surrey},%Department and Organization
            % addressline={}, 
            % city={},
            % postcode={}, 
            % state={},
            country={UK}}
\affiliation[label2]{organization={Royal Wolverhampton Hospitals NHS Trust},%Department and Organization
            % addressline={}, 
            % city={},
            % postcode={}, 
            % state={},
            country={UK}}
\affiliation[label3]{organization={Southampton Solent University},%Department and Organization
            % addressline={}, 
            % city={},
            % postcode={}, 
            % state={},
            country={UK}}
\affiliation[label4]{organization={Department of Data Science and AI, Monash University}, 
            country={Australia}}
\affiliation[label5]{organization={AIML, Adelaide University},
            country={Australia}}

%% Abstract
\begin{abstract}
Machine learning models for medical image analysis often exhibit subgroup-dependent performance, which impacts how decisions should be allocated between automated systems and human experts under limited resources. Prior work on AI fairness and human--AI cooperation, including learning to defer (L2D) and learning to complement (L2C), typically addresses these problems in isolation.
We propose \textbf{People-Centred Medical Image Analysis} (PecMan), a framework for fairness-aware human--AI cooperative classification that jointly models subgroup-dependent reliability, decision allocation, and collaborative prediction. PecMan combines subgroup-specialised predictors with a gating and consolidation mechanism that dynamically assigns cases to automated models, human experts, or their combination, without requiring sensitive attributes at test time.
We also introduce the \textbf{FairHAI} benchmark for evaluating trade-offs between predictive accuracy, subgroup equity, and human involvement. In addition, we provide a theoretical analysis of multi-agent gating via selection regret and characterise fairness--coverage trade-offs under input-dependent allocation. Experiments across multiple medical imaging datasets demonstrate that PecMan achieves consistently improved trade-offs compared to methods that address fairness or human--AI cooperation separately.
Code will be available upon paper acceptance.
\end{abstract}

%\keywords{AI Fairness  \and Learning to Defer \and People-centred AI.}

% %%Graphical abstract
% \begin{graphicalabstract}
% %\includegraphics{grabs}

% \includegraphics[width=1\linewidth]{MIA/images/highlights_1.png}

% \end{graphicalabstract}

%%Research highlights
% \begin{highlights}
% \item Unified Framework Integrating Fairness, Learning to Defer, and Learning to Complement
% \item No Need for Demographic Attributes at Test Time
% \item Strong Performance Gains Across Accuracy and Fairness
% \item Dynamic Human-AI Teaming Under Workload Constraints
% \end{highlights}
% \begin{highlights}
% \item Fairness-aware human-AI cooperation integrating learning to defer and learning to complement
% \item No need for demographic attributes at test time
% \item Improved trade-offs between diagnostic accuracy, subgroup equity, and clinician workload
% \item Benchmark protocol for evaluating fairness-aware human-AI cooperation in medical imaging
% \end{highlights}

% \begin{highlights}
% \item Fairness-aware human-AI cooperation integrating defer and complement
% \item No need for demographic attributes at test time
% \item Better trade-offs among accuracy, subgroup equity, and clinician workload
% \item Benchmark protocol for fair human-AI cooperation in medical imaging
% \item Formal analysis of multi-agent gating and fairness–coverage trade-offs
% \end{highlights}

%% Keywords

\end{frontmatter}

\section{Introduction}

Machine learning systems for pattern recognition are increasingly deployed in high-stakes domains, where predictive performance alone is insufficient to guarantee reliable decision-making. In many applications, including medical image analysis, models exhibit heterogeneous performance across data subpopulations due to dataset biases or distributional shifts. 
The rapid advancements in medical image analysis have led to highly accurate data-centric AI models that are increasingly being integrated into clinical practice~\citep{li2023medical}. These developments are particularly impactful in radiology, where the volume of medical imaging far exceeds the capacity of the available workforce~\citep{radiologyCensus2024}. By assisting with diagnostic tasks, AI has the potential to improve efficiency, reliability, and consistency in clinical workflows~\citep{topol2019high}. 
However, subgroup-dependent reliability affects not only overall model accuracy, but also how predictions should be integrated with human expertise under resource constraints. As a result, the practical use of AI in healthcare depends not only on overall predictive performance, but also on how reliably models perform across patient subgroups and how effectively they interact with clinicians in routine decision-making~\citep{abuzaid2022assessment}.

One important challenge is that medical AI models, even when highly accurate on average, may underperform for particular demographic groups because of biases in the training data or learning algorithm~\citep{oakden2020hidden}. These issues are studied in algorithmic fairness, which aims to reduce performance disparities and improve equity across diverse populations~\citep{ricci2022addressing,fang2024fairness}. 
A second challenge concerns how AI systems should be integrated into clinical workflows when clinician time is limited. This has motivated research on human--AI cooperative classification, where decisions can be made by automated models, human experts, or a combination of both. Within this setting, Learning to Defer (L2D)~\citep{madras2018predict} determines when to transfer decisions to clinicians, while Learning to Complement (L2C)~\citep{complement_wilder} seeks to combine human and AI predictions to improve joint decision-making.
Although algorithmic fairness and human--AI cooperation address complementary aspects of AI-assisted decision-making, they have largely been studied independently, limiting our understanding of how subgroup-dependent performance influences collaborative allocation and decision strategies.

\begin{figure}
    \centering
    \includegraphics[width=\linewidth]{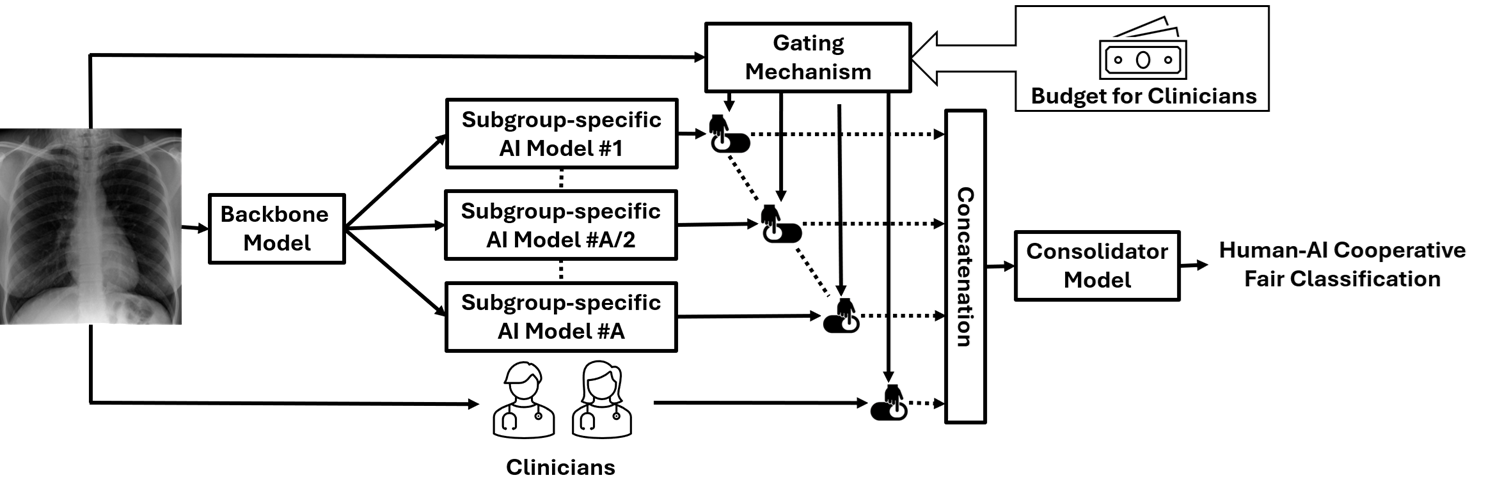}
    \caption{
    \textbf{PecMan}: A unified framework for fairness-aware human-AI cooperation. A gating mechanism selects among subgroup-specific AI models and determines whether clinician input is needed, enabling collaborative decision-making under workload constraints while improving diagnostic accuracy and subgroup equity. For visual simplicity, the framework is illustrated using generic subgroup-specific predictors; the benchmark instantiates these subgroups using dataset-specific sensitive attributes such as sex, age group, and race, as explained in Sec.~\ref{sec:benchmark}.
    }
    \label{fig:motivation}
\end{figure}

AI fairness typically seeks to reduce disparities in model performance across patient groups~\citep{zong2022medfair}, L2D focuses on optimising how responsibility is shared between clinicians and AI systems~\citep{madras2018predict}, and L2C aims to improve joint prediction by leveraging the complementary strengths of humans and machines~\citep{complement_wilder}. 

In this work, we argue that subgroup-dependent reliability provides a natural link between fairness and human-AI cooperation. Variations in model performance across subgroups influence both deferral decisions and the effectiveness of human-AI collaboration, enabling a unified analysis of trade-offs between predictive accuracy, subgroup equity, and human involvement. While some efforts have explored the intersection of fairness and L2D (e.g., Madras et al.~\citep{madras2018predict} proposed a regulariser to enforce equal false positive and false negative rates across patient subgroups within an L2D framework), these approaches typically assume prior knowledge of patient group membership at test time and do not incorporate L2C. This limits applicability in practical settings where such demographic information may be unavailable and where optimal performance may require both delegation and collaboration.

Building on this observation, we argue that \emph{(1) patient group membership should not be assumed to be available at test time}, and \emph{(2) fairness constitutes an important source of subgroup-dependent model reliability that can influence both deferral (L2D) and complementarity (L2C) decisions in human--AI collaboration}. Motivated by this perspective, we propose \textbf{Pe}ople-\textbf{C}entred \textbf{M}edical Im\textbf{a}ge A\textbf{n}alysis (PecMan), a fairness-aware extension of human-AI cooperative classification, depicted in \cref{fig:motivation}. PecMan combines subgroup-specialised AI models with a gating and consolidation mechanism to support fairness-aware deferral and complementarity under clinician workload constraints, improving diagnostic accuracy, subgroup equity, and workflow efficiency without requiring prior knowledge of patient group membership during testing.
We also propose the \textbf{Fair}ness and \textbf{H}uman-Centred \textbf{AI} (FairHAI) benchmark, designed to jointly evaluate predictive accuracy, subgroup equity, and clinician workload using public medical imaging datasets, and provide a theoretical analysis of multi-agent gating via selection regret together with a characterisation of fairness--coverage trade-offs under input-dependent allocation.
To summarise, our key contributions are:
\begin{itemize}
\item A unified framework for fairness-aware human--AI cooperative classification that integrates subgroup-specific modelling with decision deferral and complementarity;
\item A gating and consolidation mechanism that enables dynamic allocation between automated models and human experts without requiring sensitive attributes at test time;
\item A formal analysis of multi-agent gating based on selection regret and fairness--coverage trade-offs under input-dependent allocation;
\item A benchmark protocol (FairHAI) for evaluating trade-offs between predictive accuracy, subgroup equity, and human involvement;
\end{itemize}
Experimental results on the proposed FairHAI benchmark demonstrate that PecMan consistently outperforms existing approaches that treat AI fairness, L2D, and L2C in isolation, achieving improved trade-offs between predictive accuracy, subgroup equity, and human involvement. Beyond the specific application to medical imaging, these results highlight the effectiveness of modelling subgroup-dependent reliability within a unified framework for fairness-aware decision allocation, suggesting broader applicability to pattern recognition systems operating under resource constraints.

\section{Literature Review}

%This paper goes beyond classification accuracy to address the needs of key stakeholders in medical AI (i.e., clinicians, healthcare settings, patients, and regulatory agencies) by aligning system design with their workflows and priorities to foster trust, usability, and clinical effectiveness. For clinicians and healthcare providers, we advocate for AI as an assistive tool that enhances human-centred decision-making rather than replacing it~\citep{topol2019high}. For patients and regulators, we tackle critical challenges of fairness and diagnostic reliability to ensure equitable and effective care delivery~\citep{iqbal2024towards}. However, current methodologies often treat fairness and human-AI cooperation as separate objectives, overlooking their interdependence.  Addressing this gap motivates our development of PecMan, a unified framework that jointly optimises fairness and human-AI collaboration.

This section reviews prior work on AI fairness and human-AI cooperative classification in medical imaging. Our focus is on methods that address subgroup-dependent model performance and decision allocation between AI systems and clinicians under practical workload constraints. Existing approaches have typically studied fairness, learning to defer, and learning to complement separately, which motivates our development of a unified framework that addresses these objectives jointly.

\subsection{Fairness in Artificial Intelligence (AI)} 

Fairness in AI has become a central concern in healthcare due to the growing body of evidence showing that machine learning models can propagate or amplify biases related to race, gender, age, and other sensitive attributes~\citep{zong2022medfair,fang2024fairness}. 
Ensuring fair performance across diverse populations is recognised as an ethical and regulatory imperative given that algorithmic biases can undermine trust in AI systems~\citep{herington2023ethical} and exacerbate health disparities, particularly for under-represented populations~\citep{larrazabal2020gender}.

Bias mitigation strategies can be broadly categorised into pre-processing, in-processing, and post-processing techniques, as follows:
\begin{enumerate}
    \item Pre-processing approaches (e.g., SMOTE~\cite{nitesh2002smote}) re-balance datasets prior to training.
    \item Post-processing methods (e.g., equalised odds~\cite{pleiss2017fairness}) adjust outputs after training.
    \item In-processing methods directly modify the training to reduce bias while maintaining performance~\citep{madras2018learning}. 
\end{enumerate}
Among these, in-processing methods tend to show superior results, so we focus on these methods below.
For instance, adversarial training~\citep{zhang2018mitigating} has been widely used to minimise a model's ability to predict sensitive attributes, though often at the cost of reduced accuracy~\citep{madras2018learning}.
Another notable in-processing method is Group Distributionally Robust Optimisation (GroupDRO)~\cite{sagawa2019distributionally} that focuses on minimising worst-case performance across demographic groups. Although GroupDRO ensures fairness across underperforming groups, it may reduce overall model performance~\cite{zong2022medfair}.
Furthermore, domain generalization methods such as SWAD~\citep{cha2021swad} improve robustness by finding flatter loss minima, but may inadvertently prioritise majority groups and fail to optimise for group fairness.
Other in-processing approaches like representation disentanglement aim to isolate sensitive attributes from task-relevant features~\citep{tartaglione2021end, sarhan2020fairness}, though they struggle with complex, multi-source biases. More recent methods have explicitly addressed group fairness while preserving accuracy. FairSeg~\citep{tian2024fairseg} introduces Fair Error-Bound Scaling to rebalance loss across groups. FairCLIP~\citep{luo2024fairclip} applies optimal transport to align demographic distributions in vision-language models. FairVision~\citep{luo2024fairvisionequitabledeeplearning} incorporates Fair Identity Scaling to improve fairness in multimodal medical imaging.
Despite their innovation, these methods rely on a single model, often facing the challenging trade-off between group fairness and global performance. In contrast, the AI fairness part of our approach leverages multiple subgroup-specific AI models to optimise each subgroup independently, effectively balancing fairness and accuracy through an effective gating mechanism.

\subsection{Human-AI Cooperation}

Traditional machine learning systems are typically optimised without considering how they interact with human users~\citep{kamar2012combining}. Although methodologically sound, this assumption often breaks down in practice due to the complexity of real human-AI dynamics. Recent research highlights the need for AI systems that not only assist but also interact effectively with human experts~\citep{chiou2023trusting}, with factors such as model confidence~\citep{lu2021human} and explainability~\citep{weitz2019you} playing an important role in shaping that interaction. 
Crucially, this interaction must be bidirectional: while AI influences human decisions, human behaviours and judgments also shape AI outcomes. 
This duality is the core of human-AI cooperative classification (HAI-CC)~\citep{complement_wilder}, where decisions can be made in one of three ways:
\begin{enumerate}    
    \item AI predicts alone.
    \item AI defers decisions to human experts.
    \item AI makes a joint decision with human experts.
\end{enumerate}
Below, we explain the main HAI-CC approaches being studied in the literature, relating them to these three options. 

\paragraph{Learning to Defer (L2D)} L2D methods primarily focus on options (1) and (2), where either the AI or the human makes the final decision. These approaches aim to jointly learn a predictor and a rejector, enabling the AI to defer decisions to humans when it is uncertain~\citep{madras2018predict}. Early work drew on rejection learning~\citep{cortes2016learning}, and subsequent research developed surrogate loss functions that are consistent with the Bayes-optimal classifier obtained in the case of 0-1 loss~\citep{charoenphakdee2021classification, mozannar2022teaching}. 
Other advancements include score-based triage~\citep{raghu2019algorithmic} and differentiable triage~\citep{okati2021differentiable}, which assign decisions based on estimated human versus model errors. However, one common limitation of L2D-based methods is the reliance on the single-expert setting, overlooking the more complex environments with the availability of multiple experts. 
Hence, recent research in L2D has shifted the focus to the multiple-expert setting~\citep{multil2d, keswani2021towards, leitao2022human}.
Despite such an extensive research, current L2D-based learning methods have not been designed to enable AI models and human experts to jointly produce a final classification. 
To address this gap, \say{learning to complement} methods have been introduced as explained below.

\paragraph{Learning to Complement (L2C)} L2C methods explore scenarios where AI and humans combine their predictions to maximise the expected utility of the human-AI decision~\citep{complement_wilder,steyvers2022bayesian,kerrigan2021combining} (options 1 and 3). Techniques include: 1) confusion-matrix-based fusion~\citep{kerrigan2021combining}; 2) Bayesian modeling of joint predictions~\citep{steyvers2022bayesian}; and leveraging perceptual differences~\citep{liu2023humans} to improve performance through hybrid systems.
These methods show that human-AI cooperation can outperform either party alone. However, unlike L2D approaches, most L2C methods do not account for the cost of collaboration (e.g., time, effort, annotation burden), limiting practical adoption.

\paragraph{Learning to Defer and to Complement (L2D + L2C)} 
A small number of recent works explore both L2D and L2C jointly. Notably, LECODU~\citep{lecodu} integrates deferral and cooperation within a unified learning framework, marking an important step toward more comprehensive human-AI cooperative systems. 
PecMan builds on this line of work by explicitly incorporating subgroup equity into both model design and evaluation. In particular, PecMan uses subgroup-specialised predictors within a gating/consolidation framework and studies collaborative decision-making under fairness and workload constraints without requiring sensitive attributes at test time.

\subsection{Learning to Defer and AI Fairness} 

The interplay between fairness and L2D has been studied by Madras et al.\citep{madras2018predict} with the incorporation of a fairness regulariser (e.g., equalised false positives and false negatives across patient subgroups) into L2D optimisation. However, their approach requires prior knowledge of demographic group membership during testing and does not incorporate L2C. These limitations reduce applicability in real-world settings where group attributes are not guaranteed to be available, and where optimal performance may rely on joint decisions rather than deferral alone.
In contrast, our work explicitly studies how subgroup-dependent model reliability interacts with deferral and complementarity decisions, without assuming demographic group membership at test time.
By doing so, we aim to advance human-AI cooperative systems that jointly consider diagnostic accuracy, subgroup equity, and clinician workload in medical imaging settings.

\subsection{Fairness and HAI-CC Benchmarks}

Current benchmarks for HAI-CC, L2D, and L2C remain fragmented. Early L2D work introduced small-scale benchmarks using vision datasets (e.g., CIFAR), paired with simulated experts or cost-sensitive rejection models \citep{madras2018predict, charoenphakdee2021classification}, and later studies extended these evaluations to multiple-expert settings \citep{keswani2021towards}. L2C methods similarly assess complementarity on standard image classification datasets, often relying on synthetic human error models or retrospective clinician labels to estimate collaborative gains \citep{steyvers2022bayesian,kerrigan2021combining}. Although these resources provide insight into isolated aspects of collaboration, such as deferral quality or fusion benefits, they do not evaluate fairness or multi-rater variability.

In contrast, fairness in medical imaging has been supported by more structured benchmark resources. Early analyses of hidden stratification and demographic performance gaps highlighted the need for fairness-aware evaluation in clinical AI \citep{oakden2020hidden, obermeyer2019dissecting}. More recent frameworks such as MEDFAIR \citep{zong2022medfair} provide standardised protocols across multiple medical imaging modalities, enabling consistent measurement of group-specific disparities. Task-specific fairness suites, including FairVision \citep{luo2024fairvisionequitabledeeplearning} for ophthalmology and FairSeg \citep{tian2024fairseg} for segmentation, extend these ideas by incorporating fairness-aware metrics. However, these fairness benchmarks do not consider human-AI collaboration, deferral behaviour, or workflow-related constraints, leaving fairness evaluation completely decoupled from HAI-CC benchmarking. 
Consequently, existing benchmarks do not provide a unified evaluation of fairness, deferral and complementarity strategies, workload constraints, and multi-rater uncertainty in medical imaging settings, underscoring the need for more comprehensive evaluation frameworks for human-AI collaboration.

\section{Methodology}

%Our approach addresses AI fairness alongside Learning to Defer (L2D) and Learning to Complement (L2C) through a multi-objective training framework. Specifically, we train multiple subgroup-specific models and integrate them into an L2D+L2C strategy that balances fairness with effective human-AI collaboration.
Our proposed PecMan unifies AI fairness with L2D + L2C through a multi-objective framework that uses subgroup-specific models to balance equitable performance with effective human-AI collaboration.
We assume access to a dataset containing ground truth and multi-rater annotations, represented as \(\mathcal{D} = \left \{ \left (x_i,\mathcal{M}_i,a_i,y_i \right ) \right \}_{i=1}^{N}\), where \(x \in \mathcal{X} \subseteq \mathbb{R}^{H \times W \times R}\) is an  image of size $H \times W$ and $R$ colour channels, \(\mathcal{M}_i = \{\hat{y}^{m}_i\}_{m=1}^{M}\) represents a set of one-hot ``noisy'' labels, with \(\hat{y} \in \mathcal{Y} = \{ \hat{y} \in \{0, 1\}^{K}: \pmb{1}^{\top} \hat{y} = 1 \}\) denoting a ``noisy'' label with \(K\) classes, produced by $M$ annotators, \(a \in \mathcal{A} = \{ 1,\dots,A \}\) 
%denotes the demographic attribute (e.g., for the sex attribute, $\mathcal{A} = \{\text{male},\text{female}\}$),  
denotes the subgroup label associated with a sensitive attribute used for fairness-aware training (e.g., sex, age group, or race category),
and \( y \in \mathcal{Y}\) is the ground truth label. 
The inclusion of $M$ noisy labels per sample makes the framework broadly applicable to real-world clinical datasets, which often exhibit annotation variability due to differences in expertise, interpretation, and labelling protocols. 
Such a dataset definition also enables PecMan to handle both clean and noisy training sets, accommodates multi-rater scenarios, and leverages label diversity to improve robustness and fairness. By explicitly modelling annotation uncertainty, the system becomes more generalisable and better suited for deployment across heterogeneous healthcare settings.
Demographic attributes are only used during training to construct fairness-aware loss functions. During evaluation, demographic information is never used as input data; instead, we use it exclusively to compute fairness metrics and assess performance across groups.

For the sections below, assume we have: \emph{(i)} a backbone model \(f_{\theta}:\mathcal{X} \to \mathcal{F}\) that maps images to an \(F\)-dimensional feature space \(\mathcal{F} \subseteq \mathbb{R}^F\); \emph{(ii)} multiple subgroup-specific prediction models \( \{ h_{\phi_{j}}:\mathcal{F} \to \mathcal{Y} \}_{j=1}^{A}\), each taking features from the feature space \(\mathcal{F}\) to produce a probability distribution of labels in \(\mathcal{Y}\); \emph{(iii)} a gating model \(g_{\zeta}:\mathcal{X} \to [0,1]^{A+1}\) that weights the predictions by each of the \(A\) subgroup-specific models~\citep{imbert2025mixture} and the prediction by the user; and \emph{(iv)} a consolidator model \(m_{\gamma}:\mathcal{Y}^{A+1} \to \mathcal{Y}\) that takes all weighted predictions to produce a final prediction.

\subsection{Background Material}

The fairness loss function used for training our models is the Fair Identity Scaling (FIS) loss~\citep{luo2024fairvisionequitabledeeplearning} that jointly optimises overall accuracy and fairness.  For instance, to train one of the subgroup-specific models and the backbone model using a batch $\mathcal{B} \subset \mathcal{D}$, we have:
\begin{equation}
    \theta^*,\phi^* = \operatorname*{argmin}_{\theta,\phi} \frac{1}{|\mathcal{B}|}  \sum_{\left (x,\mathcal{M},a,y \right ) \in \mathcal{B}} \ell_{\mathrm{FIS}} (y, h_{\phi} \left ( f_{\theta}(x)\right), a, c, \mathcal{B} ),
    %\mathsf{w}(x, y, a, c) \times \ell_{\mathrm{BCE}} \left (h_{\phi} \left ( f_{\theta}(x)\right) ,y  \right ),
    \label{eq:optimization_step0}
\end{equation}
where
\begin{equation}
\begin{aligned}[b]
    \ell_{\mathrm{FIS}} (y, h_{\phi} \left ( f_{\theta}(x)\right), a, c, \mathcal{B} ) & =  \left[ (1-c) \times \mathsf{s}^{I}(x,y,\mathcal{B}) + c  \times \mathsf{s}^{G}(a,\mathcal{B}) \right] \\
    & \quad \times \ell_{\mathrm{BCE}} \left(y , h_{\phi} \left ( f_{\theta}(x)\right)  \right).
\end{aligned}
    \label{eq:l_fis}
\end{equation}
In \cref{eq:l_fis}, \(\ell_{\mathrm{BCE}}(.)\) denotes the binary cross-entropy loss, \(c \in [0,1]\) represents a hyper-parameter that weights the contribution between the individual-scaled loss weight of a training sample (i.e., \(\mathsf{s}^{I}(x,y)\)) and the group-scaled loss weight of a training  
sample (i.e., \(\mathsf{s}^{G}(a)\)). Both of the individual- and group-scaled weights are defined as follows:
\begin{equation}
\begin{split}
    \textstyle \mathsf{s}^{I}(x,y,\mathcal{B}) & = \frac{\exp(\ell_{\mathrm{BCE}}(h_{\phi} \left ( f_{\theta}(x)\right), y))}{\sum_{ (\tilde{x},\tilde{\mathcal{M}},\tilde{y},\tilde{a}) \in \mathcal{B}} \exp(\ell_{\mathrm{BCE}}(h_{\phi} \left ( f_{\theta}(\tilde{x})\right), \tilde{y}))},
    \\
    \quad \mathsf{s}^{G}(a,\mathcal{B}) & =  \frac{\exp\left(\mathsf{D}_{\mathrm{OT}} (\mathcal{L}(\mathcal{B}),\mathcal{L}_a(\mathcal{B})) \right)} {\sum_{j \in \mathcal{A}} \exp\left(\mathsf{D}_{\mathrm{OT}} (\mathcal{L}(\mathcal{B}),\mathcal{L}_j(\mathcal{B})) \right)},
\end{split}
\label{eq:details_step0}
\end{equation}
where \(\mathsf{D}_{\mathrm{OT}}(\mathcal{L}(\mathcal{B}), \mathcal{L}_{a}(\mathcal{B}))\) is the optimal transport divergence between the set of all losses in the batch, denoted by $\mathcal{L}(\mathcal{B})$, and the set of losses for samples of subgroup $a$ in the same batch, represented by $\mathcal{L}_a(\mathcal{B})$.
In \cref{eq:l_fis}, \( \mathsf{s}^{I}(x,y,\mathcal{B}) \) provides a larger weight to the difficult samples that contain a large loss, while \( \mathsf{s}^{G}(a,\mathcal{B}) \) provides a larger weight to the subgroups that have higher deviation in terms of group losses.

\begin{figure}[t]
    \centering
    \includegraphics[width=1\linewidth]{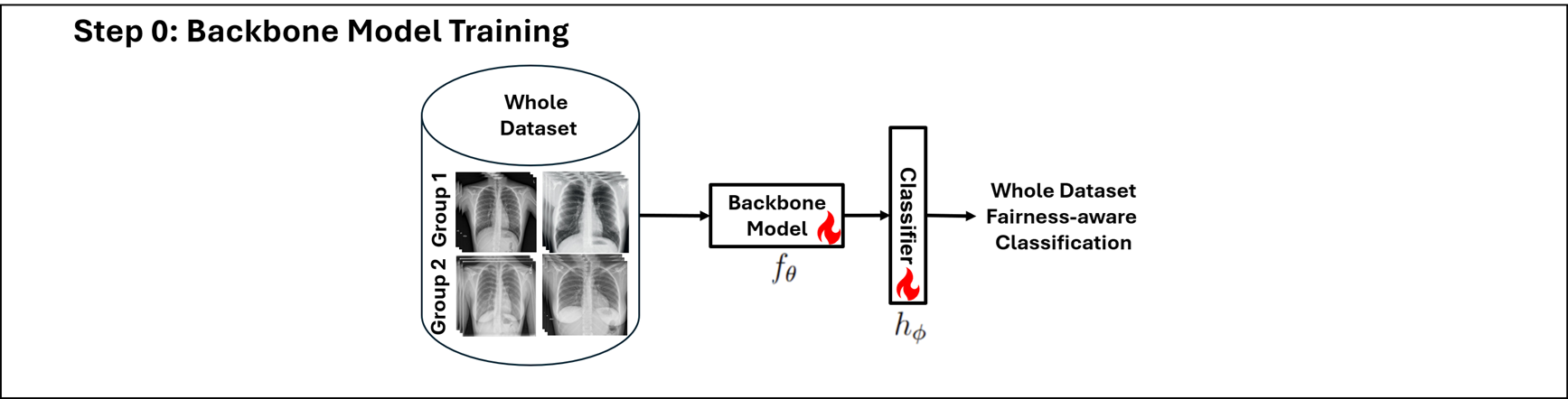}
    \caption{
    %\textbf{Step 0 – Backbone Training:} PecMan initialises its backbone model using the FIS loss~\citep{luo2024fairvisionequitabledeeplearning}, which jointly optimises overall classification accuracy and fairness across patient groups, which in this case represent the sensitive attribute \textit{sex} with values ``male'' and ``female''.
    \textbf{Step 0 -- Backbone Training:} PecMan initialises its backbone model using the FIS loss~\citep{luo2024fairvisionequitabledeeplearning}, which jointly optimises overall classification accuracy and fairness across patient subgroups. The specific subgroup definition is dataset-dependent and is described in Section~\ref{sec:benchmark}.}
    \label{fig:workflow_step0}
\end{figure}

\begin{figure}[t]
    \centering
    \includegraphics[width=1\linewidth]{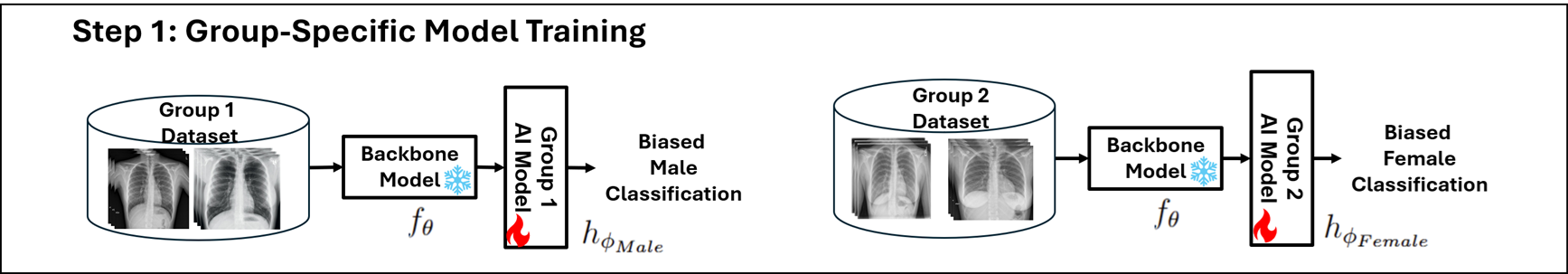}
    \caption{
    %\textbf{Step 1 – Group-specific Model Training:} This step focuses on training classifiers tailored to individual patient subgroups, enabling fairness-aware performance across demographic groups.
    \textbf{Step 1 -- Subgroup-Specific Model Training:} This step trains classifiers tailored to individual patient subgroups, enabling fairness-aware performance across subgroup-specific populations.}
    \label{fig:workflow_step1}
\end{figure}

\begin{figure}[t]
    \centering
    \includegraphics[width=1\linewidth]{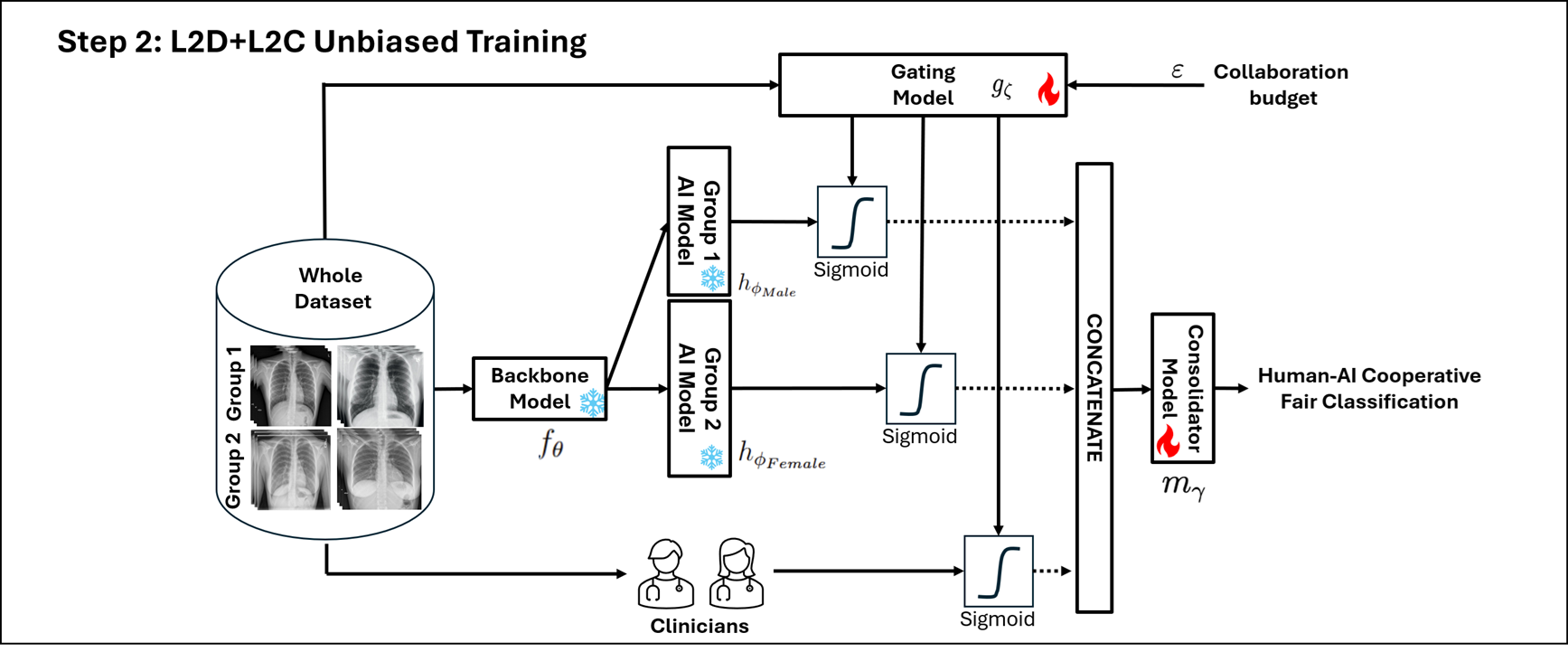}
    \caption{\textbf{Step 2 – L2D+L2C Unbiased Training:} PecMan trains the gating and consolidator models using the FIS loss, enabling unbiased decision-making that combines L2D and L2C strategies.}
    \label{fig:workflow_step2}
\end{figure}

\subsection{Training}

The training of PecMan consists of three sequential steps, as shown in \cref{fig:workflow_step0,fig:workflow_step1,fig:workflow_step2} and detailed below. 

\subsubsection{Step 0: Backbone Model Training - \cref{fig:workflow_step0}} 
We first train the backbone model \( f_{\theta^*} \) with the optimisation in \cref{eq:optimization_step0} on multiple batches \(\mathcal{B}\) sampled from the dataset \(\mathcal{D}\) using \(c=0.5\). 
This training optimises overall accuracy and fairness among patient groups to build a backbone model that serves  as the foundation for the next training steps.  %\cuong{The formulation of \(s^{G}\) is complicated and needs to explain.}

\subsubsection{Step 1: Subgroup-Specific Model Training - \cref{fig:workflow_step1}} Subgroup-specific models are trained by adding a randomly initialised classification layer \(h_{\phi_{j}}(.)\) to the frozen backbone model \(f_{\theta^{*}}(.)\) for each subgroup \( j \in \mathcal{A} \), with:  
\begin{equation}
\phi^{*}_{j} = \operatorname*{argmin}_{\phi_{j}} \frac{1}{|\mathcal{B}|} \sum_{\left (x,\mathcal{M},a,y \right ) \in \mathcal{B}} \delta(a = j) \times \ell_{\mathrm{FIS}} (y, h_{\phi_{j}} \left ( f_{\theta^{*}}(x)\right), a, c, \mathcal{B} ),
\label{eq:training_cohort_specific_model}
\end{equation}
where the indicator function \(\delta(a=j)\) only allows the training of samples belonging to subgroup \(a=j\),
the fairness hyper-parameter \(c\) is set to 0 to focus on individual accuracy, and \(\ell_{\mathrm{FIS}}(.)\) is defined in \cref{eq:l_fis}. 
Notice in \cref{eq:training_cohort_specific_model} that we could have used the BCE loss instead of the FIS loss with $c=0$, but
using the same loss family across steps ensures alignment with the overall fairness objective, even if fairness is not explicitly enforced in this Step 1.

This training enables each teacher model  \( h_{\phi^{*}_{j}} \left (f_{\theta^{*}}(x) \right ) \) to become specialised for the classification of samples belonging to the demographic group \(j \in \mathcal{A}\). 
%\cuong{For the FIS loss, when setting c = 0, it becomes a weighted binary cross-entropy loss, where the weight is the softmax on the loss values. Why is such a complicated loss like that used to train bias teachers? Why isn't the binary cross-entropy loss used to train bias teachers?}\gustavo{Good question!  We can try that too.}

\subsubsection{Step 2: L2D+L2C Unbiased Training - \cref{fig:workflow_step2}}  
This step optimises a gating model \(g_{\zeta}(.) \), which outputs the probability of selecting the type of cooperative classification (one or more of the subgroup-specific AI models, a user, or a combined prediction between AI models and a user); and a consolidator model \(m_\gamma (.)\) that takes the output of the gating model to produce a final prediction.  The training is defined by
\begin{equation}
\begin{split}
    \textstyle \zeta^*,\gamma^* = &  \operatorname*{argmin}_{\zeta,\gamma} \frac{1}{|\mathcal{B}|} \sum_{(x,\mathcal{M},y,a) \in \mathcal{B}} \ell_{\mathrm{FIS}} \left( y, \mathsf{m} \left( \gamma,\zeta;\{\phi^{*}_{j}\}_{j=1}^{A},\theta^{*},x, \hat{y} \right), a, c, \mathcal{B} \right) \\
    & \text{subject to: } \frac{1}{|\mathcal{D}|} \sum_{(x,\mathcal{M}, y, a) \in \mathcal{D}} \sum_{j \in \mathcal{A}} g^{(j)}_{\zeta}(\mathbf{x}) \ge \varepsilon,
    \label{eq:coverage_constraint}
\end{split}
\end{equation}
where \(\ell_{\mathrm{FIS}} (.)\) is defined in \cref{eq:l_fis} with \(c=0.5\) (representing a balanced overall accuracy and fairness among patient groups),
\(\varepsilon \in [0,1] \) represents the collaboration budget that is defined by the minimum proportion of cases to be analysed by the AI models, and 
%\textcolor{red}{this budget constraint doesn't work as the fact that it was analized by the AI does not mean it wasn't analyzed by humans -- shouldn't it be something like: \( \text{s.t.: } \frac{1}{|\mathcal{D}|} \sum_{(x,\mathcal{M}, y, a) \in \mathcal{D}} g^{(A+1)}_{\zeta}(\mathbf{x}) \le \varepsilon \)?}, and
\begin{equation}
\begin{aligned}[b]
    \mathsf{m} \left( \gamma,\zeta;\{\phi^{*}_{j}\}_{j=1}^{A},\theta^{*},x, \hat{y} \right)  = m_{\gamma} \big( & g^{(1)}_{\zeta}(x)\times h_{\phi^{*}_{1}}(f_{\theta^{*}}(x)), \dots, \\ 
    & g^{(A)}_{\zeta}(x)\times h_{\phi^{*}_{A}}(f_{\theta^{*}}(x)), g^{(A+1)}_{\zeta}(x)\times \hat{y} \big),
\end{aligned}
\end{equation}
denotes the prediction produced by the consolidator model that takes as inputs the group-specific model predictions and one of the user predictions \(\hat{y} \sim \mathcal{M}\),  
%\textcolor{red}{we need to tell how we sample this.} \zheng{Here, we only simulate one human expert for each dataset. In optimam, we select the majority of all human expert groups.}, 
all of them weighted by the gating model \(g^{(j)}_{\zeta}(x)\) with \(j\) representing the output index (\(j \in \{1,\dots,A\} \) indexing the predictions by the subgroup-specific models and \(j = A+1\) indexing the user prediction).

\subsection{Testing}

For a test image \( x \in \mathcal{X} \), testing produces a prediction with
\begin{equation}
\begin{split}
    m_{\gamma^{*}}\Big( & \mathsf{round}\left(g^{(1)}_{\zeta^{*}}(x)\right) \times h_{\phi^{*}_{1}}(f_{\theta^{*}}(x)), \cdots,  
    \mathsf{round}\left(g^{(A)}_{\zeta^{*}}(x) \right) \times h_{\phi^{*}_{A}}(f_{\theta^{*}}(x)), \\
    & \mathsf{round}\left(g^{(A+1)}_{\zeta^{*}}(x)\right) \times \hat{y} \Big),
\end{split}
    \label{eq:testing}
\end{equation}
where we need to use the operator \(\mathsf{round}(.)\) because the model needs to commit to a binary decision (i.e., either to use or not to use the subgroup-specific models or the users), and \( \hat{y} \sim \mathcal{M} \) denotes one of the  potentially noisy labels from annotators. Note that in this testing process, we can have an ensemble of multiple subgroup-specific models, which in turn can be combined with the user annotation if \( \mathsf{round}\left( g^{(A+1)}_{\zeta^{*}}(x)\right)  = 1  \).

\subsection{Theoretical Insights on Gating and Fairness--Coverage Trade-offs}

We present a theoretical analysis of PecMan, focusing on (i) the impact of the gating mechanism on predictive performance through selection error, and (ii) its effect on fairness under varying levels of clinician involvement. We first formalise the problem via conditional risks, then analyse how multi-agent gating reduces selection error relative to single-agent selection. Finally, we characterise the induced fairness--coverage trade-offs under input-dependent allocation.

\paragraph{Problem formulation}
Let $(x,y,a) \sim \mathcal{D}$ denote an input, label, and subgroup attribute $a \in \mathcal{A}$. Consider predictors $j \in \{1,\dots,A+1\}$ (subgroup-specific models and one human expert), each with conditional risk
\begin{equation}
L_j(x) := \mathbb{E}_{y|x}\left[\ell(y,\hat{y}_j(x))\right],
\label{eq:conditional_risk}
\end{equation}
where $\ell(\cdot,\cdot)$ is a standard loss function. The gating model outputs $g_{\zeta}(x) \in [0,1]^{A+1}$, where $g_{\zeta}^{(j)}(x)$ denotes the contribution of predictor $j$, defining a selection-and-fusion mechanism. Ignoring the consolidator, we consider the surrogate risk
\begin{equation}
\mathcal{R}(g_{\zeta}) = \mathbb{E}_x \left[ \sum_{j} g_{\zeta}^{(j)}(x) L_j(x) \right].
\end{equation}

\paragraph{Optimal gating and selection error}
With oracle risks, the optimal decision is $j^*(x)=\arg\min_j L_j(x)$, yielding single-agent gating $g_{\zeta}^{(j)*}(x)=\mathbf{1}[j=j^*(x)]$. In practice, estimated risks introduce selection error: choosing $\hat{j}(x)\neq j^*(x)$ incurs excess loss
\begin{equation}
\mathcal{E}_{\text{single}}(x) = L_{\hat{j}(x)}(x) - L_{j^*(x)}(x),
\end{equation}
which can be significant when predictors are difficult to distinguish due to noisy estimates.

% PecMan mitigates this via multi-agent gating. Assuming the consolidator satisfies
% \begin{equation}
% \ell(y,\hat{y}(x)) \le \sum_{j} \tilde{g}_{\zeta}^{(j)}(x)\,\ell(y,\hat{y}_j(x)) + \delta(x),
% \end{equation}
% with $\tilde{g}_{\zeta}^{(j)}(x)$ normalised weights and $\delta(x)\ge0$, we obtain
% \begin{equation}
% L_{\text{multi}}(x) \le \sum_{j} \tilde{g}_{\zeta}^{(j)}(x)\,L_j(x) + \delta(x),
% \end{equation}
% and hence
% \begin{equation}
% \mathcal{E}_{\text{multi}}(x)
% \le \sum_{j} \tilde{g}_{\zeta}^{(j)}(x)\big(L_j(x)-L_{j^*(x)}(x)\big) + \delta(x).
% \end{equation}
% Thus, multi-agent gating replaces worst-case selection error with a weighted aggregation of suboptimality gaps, reducing sensitivity to misallocation and improving stability when predictor risks are similar.

%\paragraph{Multi-agent regret bound} We formalise the advantage of multi-agent gating via the following result.

\noindent\textbf{Proposition 1 (Multi-agent gating reduces selection regret).}
Assume:
(i) the consolidator satisfies $\ell(y,\hat{y}(x)) \le \sum_{j} \tilde{g}_{\zeta}^{(j)}(x)\,\ell(y,\hat{y}_j(x)) + \delta(x),$
with $\tilde{g}_{\zeta}^{(j)}(x) \ge 0$, $\sum_j \tilde{g}_{\zeta}^{(j)}(x)=1$, and $\delta(x)\ge 0$;
(ii) $L_j(x)=\mathbb{E}_{y|x}[\ell(y,\hat{y}_j(x))]$ is finite for all $j$.

Then, for any input $x$, the multi-agent selection regret satisfies
\begin{equation}
\mathcal{E}_{\mathrm{multi}}(x)
\le \sum_{j} \tilde{g}_{\zeta}^{(j)}(x)\big(L_j(x)-L_{j^*(x)}(x)\big) + \delta(x),
\end{equation}
where $j^*(x)=\arg\min_j L_j(x)$.

\textit{Proof sketch.} Taking expectation over $y|x$ in (i) yields 
$L_{\mathrm{multi}}(x) \le \sum_j \tilde{g}_{\zeta}^{(j)}(x)L_j(x) + \delta(x)$.
Subtracting $L_{j^*(x)}(x)$ gives the result. Please see proof in Appendix A.

This bound shows that multi-agent gating replaces worst-case selection error with a weighted aggregation of suboptimality gaps. While it does not guarantee lower regret for every input, it reduces sensitivity to misallocation and yields lower expected regret when there is uncertainty in predictor selection.

% \paragraph{Coverage and subgroup risk}
% Let $\epsilon \in [0,1]$ denote the AI coverage, and define the subgroup risk
% \begin{equation}
% R_a(\epsilon) = \mathbb{E}\big[\ell(y,\hat{y}(x)) \mid a\big].
% \end{equation}

% \paragraph{Fairness--coverage trade-off}
% Define the fairness gap
% \begin{equation}
% \Delta(\epsilon) = \max_{a,a'} \left|R_a(\epsilon) - R_{a'}(\epsilon)\right|.
% \end{equation}
% The value of $\Delta(\epsilon)$ depends on how decisions are allocated between AI and human predictors; increasing coverage can either increase or decrease disparities depending on relative subgroup performance.

% \paragraph{Effect of gating on fairness}
% Under input-dependent allocation,
% \begin{equation}
% R_a(\epsilon) = \mathbb{E}_{x|a} \left[ \sum_{j} g_{\zeta}^{(j)}(x) L_j(x) \right]
% \approx \mathbb{E}_{x|a} \left[ \min_j L_j(x) \right].
% \end{equation}
% Thus, subgroup performance is governed by the best available predictor per input, yielding
% \begin{equation}
% \Delta(\epsilon) \le \max_{a,a'} \left| 
% \mathbb{E}_{x|a}[\min_j L_j(x)] -
% \mathbb{E}_{x|a'}[\min_j L_j(x)]
% \right|.
% \end{equation}
% Accordingly,
% \begin{equation}
% R(\epsilon) \approx \mathbb{E}_x \left[ \min_j L_j(x) \right],
% \end{equation}
% corresponding to the lower envelope of predictor risks. This formalises PecMan’s improved accuracy--coverage trade-offs via reduced selection error and subgroup-adaptive allocation.

\textbf{Proposition 2 (Fairness under input-dependent allocation).} Define the subgroup risk under gating $g_{\zeta}$ as $R_a(\epsilon) := \mathbb{E}_{x|a} \left[ \sum_{j=1}^{A+1} g_{\zeta}^{(j)}(x)\,L_j(x) \right]$, where
$L_j(x)$ is defined in \cref{eq:conditional_risk}, and the fairness gap $\Delta(\epsilon) := \max_{a,a'} \left| R_a(\epsilon) - R_{a'}(\epsilon) \right|$.
Assume that for all $x$, the gating mechanism satisfies
\begin{equation}
\sum_{j} g_{\zeta}^{(j)}(x)\,L_j(x) \le \min_j L_j(x) + \varepsilon(x),
\end{equation}
where $\varepsilon(x) \ge 0$ denotes allocation error. Then the subgroup risk satisfies
\begin{equation}
R_a(\epsilon) \le \mathbb{E}_{x|a} \left[ \min_j L_j(x) \right] + \mathbb{E}_{x|a}[\varepsilon(x)],
\end{equation}
and the fairness gap admits the bound
\begin{equation}
\Delta(\epsilon) \le \max_{a,a'} \left| 
\mathbb{E}_{x|a}[\min_j L_j(x)] -
\mathbb{E}_{x|a'}[\min_j L_j(x)]
\right| + \bar{\varepsilon},
\end{equation}
where $\bar{\varepsilon} := \max_a \mathbb{E}_{x|a}[\varepsilon(x)]$.

\textit{Proof sketch.}
Taking expectation over $x|a$ in the allocation inequality yields the bound on $R_a(\epsilon)$. The result for $\Delta(\epsilon)$ follows by applying the triangle inequality across subgroups. Please see proof in Appendix B.

This proposition shows that input-dependent gating reduces fairness disparities by approximating the subgroup-wise optimal predictor $\min_j L_j(x)$, up to an allocation error $\varepsilon(x)$. In contrast, fixed (input-independent) allocation corresponds to selecting a single predictor $j_0$, yielding subgroup risk $R_a^{\text{fixed}} = \mathbb{E}_{x|a}[L_{j_0}(x)]$. Since $\min_j L_j(x) \le L_{j_0}(x)$ for all $x$, input-dependent allocation can achieve lower subgroup risk by adapting to instance-level predictor reliability. Thus, fairness improvements arise from tailoring the allocation to subgroup-dependent predictor performance.

\section{FairHAI Benchmark} 
\label{sec:benchmark}

%In this section, we introduce \emph{FairHAI}, a benchmark designed to jointly evaluate AI fairness and HAI-CC using four public medical imaging datasets: HAM10000~\citep{tschandl2018ham10000}, CMMD~\citep{cai2023online}, CheXpert~\citep{irvin2019chexpert}, and MIMIC--CXR~\citep{johnson2019mimic}. 

%In this section, we introduce \emph{FairHAI}, a benchmark protocol for jointly evaluating fairness and human-AI cooperative classification on public medical imaging datasets (HAM10000~\citep{tschandl2018ham10000}, CMMD~\citep{cai2023online}, CheXpert~\citep{irvin2019chexpert}, and MIMIC--CXR~\citep{johnson2019mimic}) under varying levels of clinician involvement.
In this section, we introduce \emph{FairHAI}, a benchmark protocol for controlled evaluation of fairness and human-AI cooperative classification on public medical imaging datasets (HAM10000~\citep{tschandl2018ham10000}, CMMD~\citep{cai2023online}, CheXpert~\citep{irvin2019chexpert}, and MIMIC--CXR~\citep{johnson2019mimic}) under varying levels of clinician involvement.
To enable systematic comparison across competing methods, FairHAI incorporates simulated experts grounded in empirical evidence on the performance gap between clinicians and AI systems~\citep{kuo2022artificial}. The benchmark evaluates both accuracy and fairness as functions of human involvement, allowing a unified assessment of how collaborative decision-making impacts overall performance and equity.

\subsection{Datasets}  
To ensure consistency, we follow the preprocessing strategy, proposed in~\citep{zong2022medfair}, by binarising both sensitive attributes and classification labels.

\paragraph{\href{https://dataverse.harvard.edu/dataset.xhtml?persistentId=doi:10.7910/DVN/DBW86T}{HAM10000}} The dataset consists of 9,948 dermatoscopic images across seven diagnostic categories. Following~\cite{maron2019systematic}, these are grouped into \textit{benign} vs. \textit{malignant} classes. The sensitive attribute is \textit{Sex} (Male/Female), enabling fairness analysis across subgroups. The dataset includes 5,400 male images (4,492 benign, 908 malignant) and 4,548 female images (4,018 benign, 530 malignant). 

\paragraph{\href{https://wiki.cancerimagingarchive.net/pages/viewpage.action?pageId=70230508}{CMMD}} The dataset contains 5,152 mammography images annotated with diagnostic labels and a binarised age attribute. We use \textit{non-cancer} vs. \textit{cancer} classes and \textit{Age} ($\le$55 vs. $>$55) as the sensitive attribute. Malignant cases (2,614) are treated as positive, and benign + normal cases (2,538) as negative. Age distribution: i) Age $\le$55: 1,824 malignant vs. 1,972 non-malignant; and ii) Age $>$55: 790 malignant vs. 566 non-malignant. This demographic variability supports fairness analysis in our evaluation pipeline. 

\paragraph{\href{https://stanfordmlgroup.github.io/competitions/chexpert/}{CheXpert}} The dataset comprises 224,316 chest radiographs from 65,240 patients. We use \textit{No Finding} vs. \textit{Any Finding} as classes and \textit{Sex} (Male/Female) as the sensitive attribute. After filtering, the dataset includes 132,205 Male images (13,062 healthy, 119,143 unhealthy) and 90,588 Female images (9,177 healthy, 81,411 unhealthy). 

\paragraph{\href{https://physionet.org/content/mimic-cxr/}{MIMIC-CXR}} This is a large chest radiograph dataset linked to MIMIC-IV metadata. We use \textit{No Finding} vs. \textit{Any Finding} as classes and \textit{Race} (White vs. Non-White) as the sensitive attribute. After merging race data and filtering, the dataset includes 224,596 White images (78,193 healthy, 146,403 unhealthy) and 146,359 Non-White images (70,211 healthy, 76,148 unhealthy). 

\subsection{Simulation of Clinicians' Annotations to Assess HAI-CC.}
%To enable systematic comparison across methods, FairHAI simulates clinicians' annotations by assuming an error rate approximately 3\% \emph{lower} than that of the Step~0 backbone AI model, consistent with empirical observations of clinician-AI performance gaps in external validation~\citep{kuo2022artificial}. Concretely, we instantiate subgroup-specific expert accuracies as follows: 98\% for HAM10000, 95\% for CheXpert, 95\% for MIMIC--CXR, and, for CMMD, 92\% for Group~1 (age $\leq 55$) and 98\% for Group~2 (age $>55$). The injected errors follow a \emph{symmetric label noise} model, i.e., labels are uniformly flipped among incorrect classes according to the prescribed error probability, representing a standard, analytically tractable assumption in noisy-label learning~\citep{carneiro_machine_2024}.

To enable systematic comparison across methods, FairHAI simulates clinicians’ annotations by assuming an error rate approximately 3\% lower than that of the Step~0 backbone AI model, consistent with empirical observations of clinician-AI performance gaps in external validation~\citep{kuo2022artificial}. Concretely, we instantiate cohort-specific expert accuracies as follows: 98\% for HAM10000, 95\% for CheXpert, 95\% for MIMIC--CXR, and, for CMMD, 92\% for Group~1 (age $\leq 55$) and 98\% for Group~2 (age $>55$). The injected errors follow a symmetric label-noise model, i.e., labels are uniformly flipped among incorrect classes according to the prescribed error probability, representing a standard, analytically tractable assumption in noisy-label learning~\citep{carneiro_machine_2024}.

This simulation should be interpreted as providing a controlled and reproducible benchmark setting for comparing human-AI cooperation methods under varying levels of clinician involvement, rather than as a complete model of real clinical behaviour. The chosen expert accuracies are intended to reflect a plausible but simplified gap between expert and AI performance, grounded in the empirical literature cited above, while remaining consistent across datasets for systematic evaluation. Likewise, the symmetric-noise assumption is a methodological simplification that enables transparent benchmarking, and should not be interpreted as a claim that clinician errors are uniformly random or independent of case difficulty, subgroup, or workflow context. 
%Future work should therefore complement this benchmark with evaluations involving real clinical readers and more structured models of expert error.

% \paragraph{Evaluation:} binary classification is assessed with AUC, while Equity-Scaled AUC (ES-AUC)~\citep{luo2024fairvisionequitabledeeplearning} measures performance disparities across demographic subgroups to evaluate fairness. 
% Both AUC and ES-AUC are computed as a function of coverage measured on the testing set, where coverage denotes the percentage of samples classified by the AI model alone. Thus, a result with $100\%$ coverage represents the classification performed exclusively by the AI model, while $0\%$ coverage denotes a classification exclusively done by clinicians. Additionally, we provide two summarizing quantitative metrics from these accuracy-coverage curves, which show the \emph{area under the AUC-coverage curve} (AUACC) and \emph{area under the ES-AUC-coverage curve} (AUESACC), where higher AUACC and AUESACC values denote superior AUC/ES-AUC-coverage trade-offs. 
% \textcolor{red}{Statistical test?}

\subsection{Evaluation Metrics}
Binary classification performance is measured using AUC, which quantifies how well a model separates positive from negative cases, while fairness is assessed using Equity-Scaled AUC (ES-AUC)~\citep{luo2024fairvisionequitabledeeplearning}, which adjusts AUC to penalise disparities in performance across demographic groups. Both metrics are evaluated as functions of \emph{coverage}, defined as the proportion of cases handled exclusively by the AI system. In this setting, $100\%$ coverage corresponds to predictions made solely by the AI model, whereas $0\%$ coverage reflects predictions made exclusively by clinicians. To capture overall behaviour across coverage levels, we report the \emph{area under the AUC-coverage curve} (AUACC) and the \emph{area under the ES-AUC-coverage curve} (AUESACC). Higher values of these summary metrics indicate more favourable accuracy-coverage and fairness-coverage trade-offs. Confidence intervals are computed using 2{,}000 bootstrap replicates, and statistical significance is evaluated via paired one-sided t-tests.

\subsection{Baselines}
\label{sec:baselines}

The simplest baseline is the standard Expected Risk Minimisation (ERM) classifier, which trains a single model to maximise average accuracy without considering fairness or human-AI interaction.

For the L2D setting, we report results for Dependent Cross-Entropy (DCE)~\citep{dce}, a leading one-stage L2D method. We also include Fair L2D~\citep{madras2018predict}, implemented by replacing the subgroup-specific models with a single classifier trained using cross-entropy loss alongside a fairness-regularised objective. LECODU~\citep{lecodu}, which jointly models L2D and L2C, is also considered; it selects a random subgroup-specific model to combine with the simulated human prediction or defers entirely to either the human or the model.

For AI fairness comparisons, we evaluate several state-of-the-art methods, including GroupDRO~\citep{sagawa2019distributionally}, SWAD~\citep{cha2021swad}, FIS~\citep{luo2024fairvisionequitabledeeplearning}, and FairDi~\citep{masroor2024fair}. 
Because these fairness-only baselines do not incorporate clinician collaboration, they produce a single operating point at 100\% coverage; to compute AUACC and AUESACC, we pair this AI-only point with the human-only performance at 0\% coverage. 
We also propose L2D-FairDi, a simple hybrid method that combines L2D with FairDi by selecting either the Step~2 FairDi model or the simulated expert for prediction.

\section{Experiments}
\label{sec:experiments}

In this section, we first describe the training setup used for all models, followed by the results of the FairHAI benchmark. We then analyse PecMan's deferral behaviour to subgroup-specific models and clinicians, and conclude with an ablation study evaluating the contribution of the subgroup-specific models.

\subsection{Training Setup}
\label{sec:training_setup}

Our experiments are conducted on a computer using one NVIDIA RTX A6000 GPU. The implementation is built on Python $3.12.7$ and PyTorch $2.2.2$.
For the backbone model of Steps 0 and 1, we use MVCCL~\citep{chen2022multi} for CMMD mammogram dataset and ResNet18~\citep{he2016deep} for the others (following the experimental setting in~\citep{zong2022medfair}). ResNet18~\citep{he2016deep} is utilised for Step 2 in all datasets.
For the optimiser, in Step 0, we use Adam and an initial learning rate of $1\times10^{-4}$ that is decayed by a factor of $0.1$ every $10$ epochs, for a total of $30$ epochs. In Steps 1 and 2, we use SGD with a momentum of 0.9 and a weight decay of \(5 \times 10^{-4}\). For the training of subgroup-specific models, we fine-tune the final classification layer of the backbone with a learning rate of $1\times10^{-4}$. 
For the gating model trained in Step 2, the learning rate is set at 0.01 for a total of 60 epochs.

\begin{figure}[t]
    \centering
    \hspace{3em}\input{legend}\\

    \vspace{-3ex}
    \subfloat{
        % \label{fig:ham_gender_auc}
        \input{ham_gender_auc}
    }
    \hspace*{-1em}
    \subfloat{
        % \label{fig:CMMD_auc}
        \input{CMMD_auc}
    }
    \hspace*{-1em}
    \subfloat{
        % \label{fig:cxp_auc}
        \input{CXP_auc}
    }
    \hspace*{-1em}
    \subfloat{
        % \label{fig:mimic_auc}
        \input{MIMIC_auc}
    }
    \\

    \setcounter{subfigure}{0}

    \vspace{-2em}\hspace*{-1.4ex}
    \subfloat[HAM10000.]{
        \label{fig:ham_gender_esauc}
        \input{ham_gender_es_auc}
    }
    \hspace*{-3ex}
    \subfloat[CMMD.]{
        \label{fig:CMMD_esauc}
        \input{CMMD_esauc}
    }
    \hspace*{-3ex}
    \subfloat[CheXpert.]{
        \label{fig:cxp_esauc}
        \input{CXP_esauc}
    }
    \hspace*{-3ex}
    \subfloat[MIMIC.]{
        \label{fig:mimic_esauc}
        \input{MIMIC_esauc}
    }
    \caption{The AUC vs coverage (top row) and ES-AUC vs. coverage (bottom row) of competing methods on FairHAI benchmark, where each column corresponds to one dataset.}
    \label{fig:sexp}
\end{figure}

\subsection{Results}

\begin{table}[t]
    \centering
    \caption{Comparison of AUAUC and AUESAUC between different methods evaluated on four datasets.}
    \label{tab:auc}
    \resizebox{1.\linewidth}{!}{
        \begin{tabular}{l c c c c c c c c c c c c}
            \toprule
            \multirow{2}{*}{} & \multirow{2}{*}{\rotatebox{-30}{\bfseries Fairness}} & \multirow{2}{*}{\rotatebox{-30}{\bfseries L2D}} & \multirow{2}{*}{\rotatebox{-30}{\bfseries L2C}} & \multicolumn{4}{c}{\bfseries AUAUC (\%) \(\uparrow\)} & \multicolumn{4}{c}{\bfseries AUESACC (\%) \(\uparrow\)} \\ 
            \cmidrule(lr){5-8} \cmidrule(lr){9-12} 
             & & & & \bfseries HAM10000 & \bfseries CMMD & \bfseries CheXpert & \bfseries MIMIC-CXR& \bfseries HAM10000 & \bfseries CMMD & \bfseries CheXpert & \bfseries MIMIC\\
            \midrule
            ERM & \textcolor{BrickRed}{\ding{55}} & \textcolor{BrickRed}{\ding{55}} & \textcolor{BrickRed}{\ding{55}} & $91.33_{\pm 0.68}$ & $88.28_{\pm 0.31}$ & $91.21_{\pm 0.10}$ & $90.51_{\pm 0.08}$ & $90.48_{\pm 0.95}$ & $84.24_{\pm 0.83}$ & $91.56_{\pm 0.31}$ & $90.27_{\pm 0.08}$\\
            GroupDRO & \textcolor{PineGreen}{\checkmark} & \textcolor{BrickRed}{\ding{55}} & \textcolor{BrickRed}{\ding{55}} & $92.25_{\pm 0.51}$ & $84.42_{\pm 0.36}$ & $90.71_{\pm 0.59}$ & $90.20_{\pm 0.16}$ & $91.70_{\pm 0.40}$ & $80.49_{\pm 0.41}$ & $90.64_{\pm 0.83}$ & $89.97_{\pm 0.27}$ \\
            SWAD & \textcolor{PineGreen}{\checkmark} & \textcolor{BrickRed}{\ding{55}} & \textcolor{BrickRed}{\ding{55}} & $94.45_{\pm 0.19}$ & $85.17_{\pm 0.49}$ & $91.97_{\pm 0.05}$ & $90.93_{\pm 0.02}$ & $94.20_{\pm 0.46}$ & $81.19_{\pm 0.76}$ & $91.89_{\pm 0.11}$ & $90.68_{\pm 0.07}$ \\
            FIS & \textcolor{PineGreen}{\checkmark} & \textcolor{BrickRed}{\ding{55}} & \textcolor{BrickRed}{\ding{55}} & $95.23_{\pm 0.03}$ & $90.59_{\pm 0.30}$ & $91.33_{\pm 0.04}$ & $91.31_{\pm 0.06}$ & $94.99_{\pm 0.07}$ & $87.11_{\pm 0.06}$ & $91.02_{\pm 0.06}$ & $91.04_{\pm 0.07}$ \\
            FairDi & \textcolor{PineGreen}{\checkmark} & \textcolor{BrickRed}{\ding{55}} & \textcolor{BrickRed}{\ding{55}} & $95.92_{\pm 0.04}$ & $91.03_{\pm 0.05}$ & $91.52_{\pm 0.01}$ & $91.43_{\pm 0.03}$ & $95.61_{\pm 0.08}$ & $87.57_{\pm 0.49}$ & $91.44_{\pm 0.01}$ & $91.03_{\pm 0.02}$  \\
            DCE & \textcolor{BrickRed}{\ding{55}} & \textcolor{PineGreen}{\checkmark} & \textcolor{BrickRed}{\ding{55}}  & $91.55_{\pm 0.21}$ & $86.17_{\pm 0.28}$ & $90.67_{\pm 0.13}$ & $90.73_{\pm 0.17}$ & $90.59_{\pm 0.16}$ & $82.49_{\pm 0.21}$ & $90.29_{\pm 0.13}$ & $90.55_{\pm 0.17}$ \\
            Fair L2D & \textcolor{PineGreen}{\checkmark} & \textcolor{BrickRed}{\ding{55}}  & \textcolor{BrickRed}{\ding{55}} & $93.97_{\pm 0.18}$ & $91.95_{\pm 0.12}$ & $91.66_{\pm 0.14}$ & $90.97_{\pm 0.11}$ & $93.61_{\pm 0.16}$ & $88.57_{\pm 0.23}$ & $91.47_{\pm 0.13}$ & $90.84_{\pm 0.11}$ \\
            LECODU & \textcolor{BrickRed}{\ding{55}} & \textcolor{PineGreen}{\checkmark} & \textcolor{PineGreen}{\checkmark} & $94.60_{\pm 0.15}$ & $92.12_{\pm 0.22}$ & $92.29_{\pm 0.12}$ & $91.20_{\pm 0.14}$ & $94.44_{\pm 0.15}$ & $88.63_{\pm 0.22}$ & $92.12_{\pm 0.13}$ & $90.96_{\pm 0.14}$ \\
            PecMan   & \textcolor{PineGreen}{\checkmark} & \textcolor{PineGreen}{\checkmark} & \textcolor{PineGreen}{\checkmark}  & {\boldmath $97.63_{\pm 0.14}$} & {\boldmath $92.91_{\pm 0.13}$} & {\boldmath $93.77_{\pm 0.14}$} & {\boldmath $93.01_{\pm 0.10}$} & {\boldmath $97.21_{\pm 0.07}$} & {\boldmath $89.70_{\pm 0.17}$}  & {\boldmath $93.54_{\pm 0.14}$} & {\boldmath $92.22_{\pm 0.10}$} \\
            \bottomrule
        \end{tabular}
    }
\end{table}

%\cref{fig:sexp} reports the mean AUC and ES-AUC coverage curves (with standard deviation error bars) for all baselines and for PecMan, shown in the \emph{top} and \emph{bottom} rows, respectively. Across all four datasets, PecMan consistently outperforms competing methods over the full coverage range. DCE shows noticeably weaker performance than Fair~L2D, LECODU, and PecMan, particularly at higher coverage levels, likely because its unified gating-classification architecture produces an overly complex decision layer that does not generalise well on smaller datasets such as HAM10000 and CMMD. On larger datasets like CheXpert and MIMIC-CXR, however, DCE performs competitively (see \cref{tab:auc}). Fair~L2D and LECODU exhibit broadly similar performance, with LECODU showing slight advantages on HAM10000 and CheXpert. As noted in \cref{sec:baselines}, FairDi only provides results at 0\% (human-only) and 100\% (model-only) coverage. At 100\% coverage, FairDi performs strongly and generally surpasses all baselines except PecMan, which maintains a clear advantage. A notable observation is PecMan’s superior performance at 0\% coverage, achieved because clinician decisions are complemented by multiple subgroup-specific models, boosting accuracy beyond all other methods.

\cref{fig:sexp} reports the mean AUC and ES-AUC coverage curves (with standard deviation error bars) for all baselines and for PecMan, shown in the \emph{top} and \emph{bottom} rows, respectively. Across all four datasets, PecMan consistently outperforms competing methods over the full coverage range. DCE shows noticeably weaker performance than Fair~L2D, LECODU, and PecMan, particularly at higher coverage levels, likely because its unified gating-classification architecture produces an overly complex decision layer that does not generalise well on smaller datasets such as HAM10000 and CMMD. On larger datasets like CheXpert and MIMIC-CXR, however, DCE performs competitively (see \cref{tab:auc}). Fair~L2D and LECODU exhibit broadly similar performance, with LECODU showing slight advantages on HAM10000 and CheXpert. As noted in \cref{sec:baselines}, FairDi only provides results at 0\% (human-only) and 100\% (model-only) coverage. At 100\% coverage, FairDi performs strongly and generally surpasses all baselines except PecMan, which maintains a clear advantage. A notable observation is PecMan’s superior performance at 0\% coverage, achieved because clinician decisions are complemented by multiple subgroup-specific models, boosting accuracy beyond all other methods. Relative to LECODU, the main value of PecMan is therefore not only improved average performance, but also a more balanced trade-off across diagnostic accuracy, subgroup equity, and clinician involvement, which is consistent with the design goal of fairness-aware collaborative allocation.

We further quantify performance using AUACC and AUESACC, calculated from the curves in \cref{fig:sexp} and summarised in \cref{tab:auc}. PecMan achieves the highest scores on both metrics across all datasets, outperforming all baselines by substantial margins, with all differences statistically significant ($p < 0.001$).

\begin{figure}[t]
    \centering
    \hspace*{-3ex}
    \subfloat[Deferral distribution]{
        \input{deferral}
        \label{fig:deferral_distribution}
    }
    \hspace*{-1.em}
    \subfloat[Confusion matrix (\%)]{
        \input{confusion_matrix}
        \label{fig:deferral_confusion_matrix}
    }
    \hspace*{-1.75em}
    \subfloat[AUC of each component]{
        \label{fig:auc_each_component}
        \input{auc_each_component}
    }
    \caption{Performance analysis of PecMan on the testing samples of HAM10000.  (a) The deferral distribution at various clinician collaboration budgets represented by \((1-\varepsilon)\) from \cref{eq:coverage_constraint} (i.e., the proportion of cases to be analised by clinicians). (b) Confusion matrix showing the true subgroup of each deferred sample (rows) versus the subgroup-specific model or clinician it was deferred to (columns).    
    (c) Performance of different subgroup-specific model and human on the testing set of the dataset HAM10000.}
    \label{fig:deferral}
\end{figure}

% \subsection{Analysis on the deferral to subgroup-specific models and clinicians}
% We provide the deferral distribution of PecMan on the testing set of HAM10000 at various coverage rates in \cref{fig:deferral_distribution}. The results show that the gating model in PecMan can accurately defer testing samples to clinicians, according to the coverage rate shown in the x-axis. 
% To further analyse PecMan's effectiveness, we break down the number of deferred samples at the coverage of \(\varepsilon = 0.5\) and show the confusion matrix in \cref{fig:deferral_confusion_matrix}. Considering the deferral to subgroup-specific models (i.e., first two columns of the confusion matrix), PecMan makes about 14.56\% ``false'' deferral (computed from 6.88\% + 7.68\% ), while achieving 34.74\% ``true'' one (computed from 19.18\% + 15.56\% ) 
% Note that each subgroup-specific model performs optimally on samples for its own subgroup, but it also performs adequately on samples from a different subgroup, as shown in \cref{fig:auc_each_component}. Hence, PecMan can still achieve high performance overall, as shown in \cref{tab:auc}.

\subsection{Analysis of Deferral to subgroup-Specific Models and Clinicians}

\cref{fig:deferral_distribution} shows the deferral distribution of PecMan on the HAM10000 testing set across different coverage levels. As coverage varies, the gating model adjusts the proportion of cases deferred to clinicians accordingly, demonstrating effective control over human involvement.

To examine PecMan's behaviour in more detail, we analyse deferral patterns at a coverage level of $\varepsilon = 0.5$ and present the corresponding confusion matrix in \cref{fig:deferral_confusion_matrix}. Focusing on deferrals to subgroup-specific models (the first two columns), PecMan exhibits approximately 14.56\% ``wrong'' deferrals (6.88\% + 7.68\%) and 34.74\% ``correct'' deferrals (19.18\% + 15.56\%). 

Furthermore, although each subgroup-specific model achieves optimal performance on its own subgroup, it also performs adequately on samples from the other subgroup, as shown in \cref{fig:auc_each_component}. This cross-subgroup robustness contributes to PecMan's strong overall performance, reflected in the results reported in \cref{tab:auc}.

\begin{figure}[t]
    \centering
    \input{ablation_coverage_legend}
    \\
    \vspace{-1em}
    \hspace*{-1em}
    \subfloat[Male-AUC]{
        \label{fig:ham_male}
        \input{ham_male}
    }
    \hspace*{-4ex}
    \subfloat[Female-AUC]{
        \label{fig:ham_female}
        \input{ham_female}
    }
    \hspace*{-4ex}
    \subfloat[Overall-AUC]{
        \label{fig:ablation_ham_gender_auc}
        \input{ablation_ham_gender_auc}
    }
    \hspace*{-4ex}
    \subfloat[ES-AUC]{
        \label{fig:ablation_ham_gender_esauc}
        \input{ablation_ham_gender_esauc}
    }
    \caption{The subgroup-specific AUC (a,b), overall AUC (c), and ES-AUC vs. coverage curves of L2D-FairDi and PecMan on HAM10000.}
    \label{fig:ablation_l2d_fairdi}
\end{figure}

\subsection{Ablation study on the subgroup-specific models}
    We perform an ablation study on PecMan by replacing the multiple subgroup-specific models with a single fair model trained on a fairness method (i.e., FairDi), referred to as L2D-FairDi. We evaluate both L2D-FairDi and PecMan on HAM10000 and show the results in \cref{fig:ablation_l2d_fairdi}. In general, PecMan outperforms L2D-FairDi on both subgroup-specific (i.e., \cref{fig:ham_male,fig:ham_female}) and overall evaluation (i.e., \cref{fig:ablation_ham_gender_auc,fig:ablation_ham_gender_esauc}). 
    Comparing L2D-FairDi (\cref{fig:ablation_l2d_fairdi}) against standard L2D baselines (\cref{fig:sexp}) demonstrates that using a fair backbone (e.g., FairDi) can also improve the performance of L2D.

\subsection{Qualitative results}

\cref{tab:case} shows examples of PecMan’s predictions at $50\%$ coverage on HAM10000 test samples, including the input image, sensitive label (sex), subgroup-specific model predictions, clinician labels, gating model’s selection, PecMan’s aggregated prediction, and the ground-truth. Notably, when either the AI or the clinician makes a mistake, PecMan often remains correct, illustrating robustness from combining clinician decisions with subgroup-specific model outputs.

\begin{table}[t!]
\centering
\resizebox{1.\linewidth}{!}{
\begin{tabular}{lcccccc}
\toprule
Image & Sex & AI Models & Clinician Label  & \(g_{\zeta}(\mathbf{x})\) & PecMan Prediction & Ground Truth \\ 
& & [Male-AI, Female-AI] & & [Male-AI, Female-AI, Clinician] \\
\midrule
\begin{minipage}[b]{0.1\columnwidth}
	\centering
	\raisebox{-.5\height}{\includegraphics[width=\linewidth]{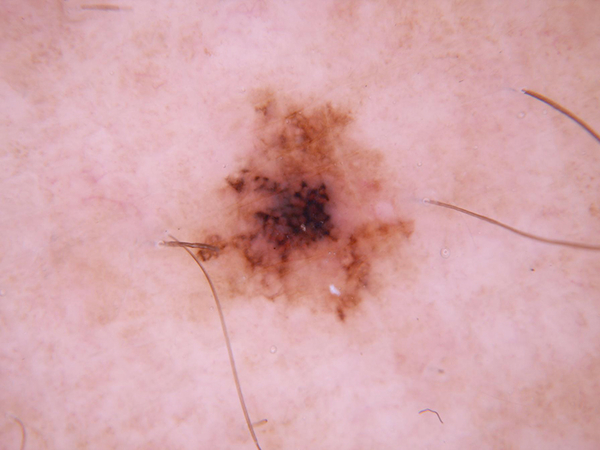}}
\end{minipage} & F & {[}Malignant, Malignant{]} & Malignant & [0.73,0.82,0.57] & Malignant & Malignant  \\
\begin{minipage}[b]{0.1\columnwidth}
	\centering
	\raisebox{-.5\height}{\includegraphics[width=\linewidth]{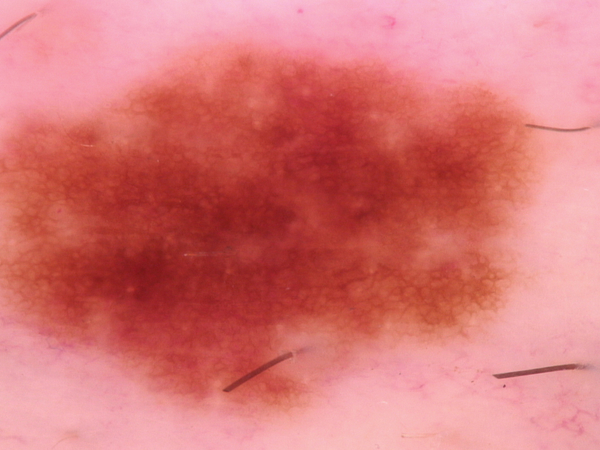}}
\end{minipage} & M & {[}Benign, Malignant{]} & Benign & [0.76,0.42,0.42] & Benign & Benign  \\
\begin{minipage}[b]{0.1\columnwidth}
	\centering
	\raisebox{-.5\height}{\includegraphics[width=\linewidth]{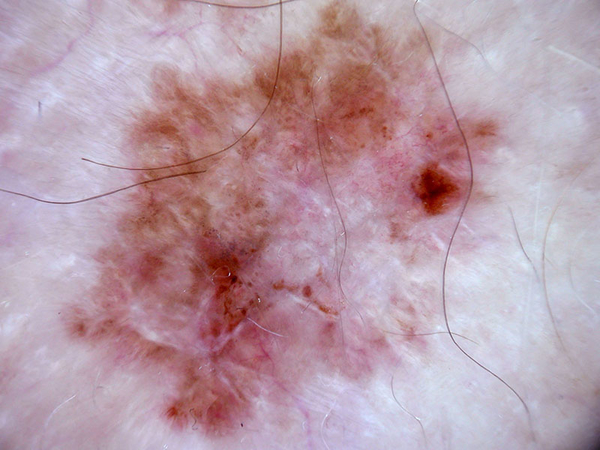}}
\end{minipage} & M & {[}Malignant, Benign{]} & Malignant & [0.63,0.77,0.53] & Malignant & Malignant  \\
\begin{minipage}[b]{0.1\columnwidth}
	\centering
	\raisebox{-.5\height}{\includegraphics[width=\linewidth]{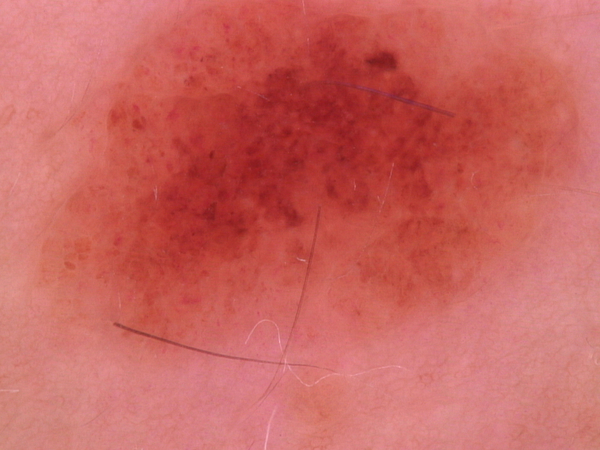}}
\end{minipage} & F & {[}Malignant, Benign{]} & Malignant & [0.21,0.85,0.61] & Benign & Benign  \\
\bottomrule 
\end{tabular}}
\caption{Case studies presenting the test image, the sensitive attribute (sex), male and female subgroup‑specific predictions,  clinician labels, the gating model’s selection, PecMan’s final prediction, and the ground truth.}
\label{tab:case}
\end{table}

\subsection{Training and testing times}

The training and inference times for all methods on the HAM10000 dataset are shown in \cref{fig:training_time,fig:inference_time}. PecMan exhibits training and testing times comparable to those of L2D/L2C-based approaches, which are generally higher than those of fairness-only methods such as GroupDRO and SWAD.

\begin{figure}[t]
    \centering
    \scalebox{0.7}{
    \begin{tikzpicture}
        \pgfplotstableread[col sep=&, row sep=\\, header=true]{
            method & time\\
            ERM & 19.75\\
            GroupDRO & 28.78\\
            SWAD & 19.12\\
            FIS & 17.56 \\
            FairDi & 40.00 \\
            DCE & 35.15\\
            Fair L2D & 35.27 \\
            LECODU & 36.72\\
            PecMan (Step 0) & 17.56\\
            PecMan (Step 1) & 5.53\\
            PecMan (Step 2) & 13.18 \\
            PecMan (Total) & 36.27 \\
        } \mytable
        % get number of rows
        \pgfplotstablegetrowsof{\mytable}
        \pgfmathsetmacro{\NumRows}{\pgfplotsretval-1}  

        %\hspace{-0.5em}
        \begin{axis}[
            width = 0.75\linewidth,
            xbar=0pt,
            y=1.5em,
            ytick=data,
            ymax={\NumRows + 1},
            ymin={-0.75},
            yticklabels from table={\mytable}{method},
            ylabel style={font=\footnotesize},
            yticklabel style = {font=\scriptsize, align=left},
            ytick pos=left,
            xticklabel style = {font=\scriptsize},
            xlabel={Time (in minutes)},
            xlabel style={font=\footnotesize},
            axis x line*=bottom,
            axis y line*=left,
            axis line style={-Latex},
            scale only axis,
            enlarge x limits=auto,
            enlarge y limits=auto,
            nodes near coords,
            every node near coord/.append style={font=\scriptsize},
            every axis plot/.append style={fill=PineGreen, draw=none}
            % every axis plot/.append style={fill=NavyBlue, draw=none}
        ]
            \addplot[] table [y expr=\NumRows - \coordindex, x=time]{\mytable};
        \end{axis}
    \end{tikzpicture}}
    \caption{Training time of PecMan and competing methods on HAM10000 dataset.}
    \label{fig:training_time}
\end{figure}

\begin{figure}[t]
    \centering
    \scalebox{0.55}{
    \begin{tikzpicture}
        \pgfplotstableread[col sep=&, row sep=\\, header=true]{
            method & time\\
            ERM & 66\\
            GroupDRO & 66\\
            SWAD & 66\\
            FIS & 66\\
            FairDi & 66\\
            DCE & 94\\
            Fair L2D & 92 \\
            LECODU & 107 \\
            PecMan & 92 \\
        } \mytable
        % get number of rows
        \pgfplotstablegetrowsof{\mytable}
        \pgfmathsetmacro{\NumRows}{\pgfplotsretval-1}  

        %\hspace{-0.5em}
        \begin{axis}[
            width = 0.75\linewidth,
            xbar=0pt,
            y=1.5em,
            ytick=data,
            ymax={\NumRows + 1},
            ymin={-0.75},
            yticklabels from table={\mytable}{method},
            yticklabel style = {font=\small, align=left},
            ytick pos=left,
            xticklabel style = {font=\small},
            xlabel={Time (in seconds)},
            axis x line*=bottom,
            axis y line*=left,
            axis line style={-Latex},
            scale only axis,
            enlarge x limits=auto,
            enlarge y limits=auto,
            nodes near coords,
            every node near coord/.append style={font=\footnotesize},
            every axis plot/.append style={fill=PineGreen, draw=none}
            % every axis plot/.append style={fill=NavyBlue, draw=none}
        ]
            \addplot[] table [y expr=\NumRows - \coordindex, x=time]{\mytable};
        \end{axis}
    \end{tikzpicture}}
    \caption{Inference time of PecMan and competing methods on HAM10000 dataset.}
    \label{fig:inference_time}
\end{figure}

\section {Discussion and Conclusion} \label{sec:conclusion}

%We presented \emph{PecMan}, a unified framework that integrates AI fairness with human-AI cooperation to improve patient equity and clinical efficacy. 
We presented \emph{PecMan}, a unified framework for fairness-aware human-AI cooperation in medical imaging under clinician workload constraints. PecMan combines subgroup-specialised AI models with a gating and consolidation mechanism to support both deferral and complementarity without requiring demographic attributes at test time, and is further supported by a formal analysis of multi-agent gating via selection regret and fairness--coverage trade-offs under input-dependent allocation. 
PecMan should be interpreted as treating fairness as one important source of heterogeneous model reliability in human-AI cooperation, rather than assuming that fairness fully defines deferral or complementarity decisions.
Beyond the method, a key contribution of this work is \emph{FairHAI}, a benchmark that jointly evaluates accuracy, fairness, and human involvement across multiple public medical imaging datasets, providing a standardised protocol for comparing HAI-CC strategies under fairness and workload considerations. 
Across the medical imaging datasets included in FairHAI, PecMan achieved improved trade-offs between diagnostic accuracy, subgroup equity, and clinician involvement compared with methods that address these components in isolation.

A practical limitation of PecMan is its reliance on subgroup-specialised models, which can improve collaborative allocation and subgroup equity but increase maintenance, monitoring, and validation complexity as the number of clinically relevant subgroups grows. In addition, the current evaluation uses binarised sensitive attributes, which supports controlled benchmarking but does not capture more complex settings involving multiple or intersectional subgroup structures. Furthermore, the FairHAI benchmark relies on simulated clinician annotations, providing a reproducible testbed but not fully reflecting real clinical behaviour, including heterogeneous expertise, biases, and human–AI interaction effects. Finally, although sensitive attributes are not required at test time, they are needed during training and evaluation, which may limit applicability when demographic metadata are incomplete or unreliable. Addressing these limitations (through more efficient subgroup modelling, evaluation with real clinical readers, and extensions to richer and noisier subgroup definitions) is an important direction for future work.

\section*{Declaration of generative AI and AI-assisted technologies in the manuscript preparation process}

During the preparation of this work the author(s) used Microsoft Copilot in order to assist with wording refinement, clarity improvements, and formatting of sections of the manuscript. The author(s) reviewed and edited the output as needed and take full responsibility for the content of the published article.

\section*{Acknowledgement}

Supported by the Engineering and Physical Sciences Research Council (EPSRC) through grant EP/Y018036/1.

% \newpage

\bibliographystyle{elsarticle-num}
\bibliography{library}

\newpage
\setcounter{page}{1}

\section*{Appendix A. Proof of Multi-Agent Regret Bound}
\label{sec:Proof of Multi-Agent Regret Bound}

\textbf{Proposition 1 (Multi-agent gating reduces selection regret).}
Let $(x,y,a) \sim \mathcal{D}$ and consider predictors $\{\hat{y}_j(x)\}_{j=1}^{A+1}$ with conditional risks
\begin{equation}
L_j(x) := \mathbb{E}_{y|x}[\ell(y,\hat{y}_j(x))],
\end{equation}
where $\ell(\cdot,\cdot)$ is a non-negative loss function. Let $j^*(x) = \arg\min_j L_j(x)$.

Assume that the consolidator $m_\gamma$ produces a prediction $\hat{y}(x)$ satisfying
\begin{equation}
\ell(y,\hat{y}(x)) \le \sum_{j=1}^{A+1} \tilde{g}_{\zeta}^{(j)}(x)\,\ell(y,\hat{y}_j(x)) + \delta(x),
\end{equation}
where $\tilde{g}_{\zeta}^{(j)}(x) \ge 0$, $\sum_j \tilde{g}_{\zeta}^{(j)}(x) = 1$, and $\delta(x) \ge 0$.

Then, the multi-agent selection regret satisfies
\begin{equation}
\mathcal{E}_{\mathrm{multi}}(x)
:= L_{\mathrm{multi}}(x) - L_{j^*(x)}(x)
\le \sum_{j=1}^{A+1} \tilde{g}_{\zeta}^{(j)}(x)\big(L_j(x) - L_{j^*(x)}(x)\big) + \delta(x),
\end{equation}
where $L_{\mathrm{multi}}(x) := \mathbb{E}_{y|x}[\ell(y,\hat{y}(x))]$.

\paragraph{Proof.}
By assumption, for any $(x,y)$,
\begin{equation}
\ell(y,\hat{y}(x)) \le \sum_{j} \tilde{g}_{\zeta}^{(j)}(x)\,\ell(y,\hat{y}_j(x)) + \delta(x).
\end{equation}
Taking expectation with respect to $y|x$ yields
\begin{equation}
L_{\mathrm{multi}}(x) = \mathbb{E}_{y|x}[\ell(y,\hat{y}(x))]
\le \sum_{j} \tilde{g}_{\zeta}^{(j)}(x)\,\mathbb{E}_{y|x}[\ell(y,\hat{y}_j(x))] + \delta(x).
\end{equation}
Using the definition of $L_j(x)$, we obtain
\begin{equation}
L_{\mathrm{multi}}(x) \le \sum_{j} \tilde{g}_{\zeta}^{(j)}(x)\,L_j(x) + \delta(x).
\end{equation}
Subtracting $L_{j^*(x)}(x)$ from both sides gives
\begin{equation}
\mathcal{E}_{\mathrm{multi}}(x)
\le \sum_{j} \tilde{g}_{\zeta}^{(j)}(x)\big(L_j(x) - L_{j^*(x)}(x)\big) + \delta(x).
\end{equation}
\hfill $\square$

\section*{Appendix B. Proof of Fairness under Input-Dependent Allocation}

\textbf{Proposition 2 (Fairness–coverage under input-dependent allocation).}
Let $(x,y,a) \sim \mathcal{D}$ and consider predictors $\{\hat{y}_j(x)\}_{j=1}^{A+1}$ with conditional risks
\begin{equation}
L_j(x) := \mathbb{E}_{y|x}[\ell(y,\hat{y}_j(x))].
\end{equation}
Define the subgroup risk under gating $g_{\zeta}$ as
\begin{equation}
R_a(\epsilon) := \mathbb{E}_{x|a} \left[ \sum_{j=1}^{A+1} g_{\zeta}^{(j)}(x)\,L_j(x) \right],
\end{equation}
and the fairness gap
\begin{equation}
\Delta(\epsilon) := \max_{a,a'} \left| R_a(\epsilon) - R_{a'}(\epsilon) \right|.
\end{equation}

Assume that the gating mechanism satisfies, for all $x$,
\begin{equation}
\sum_{j=1}^{A+1} g_{\zeta}^{(j)}(x)\,L_j(x) \le \min_{j} L_j(x) + \varepsilon(x),
\end{equation}
where $\varepsilon(x) \ge 0$ denotes the allocation error.

Then the subgroup risk is bounded by
\begin{equation}
R_a(\epsilon) \le \mathbb{E}_{x|a} \left[ \min_{j} L_j(x) \right] + \mathbb{E}_{x|a}[\varepsilon(x)],
\end{equation}
and the fairness gap satisfies
\begin{equation}
\Delta(\epsilon) \le \max_{a,a'} \left| 
\mathbb{E}_{x|a}[\min_{j} L_j(x)] -
\mathbb{E}_{x|a'}[\min_{j} L_j(x)]
\right| + \bar{\varepsilon},
\end{equation}
where
\begin{equation}
\bar{\varepsilon} := \max_{a} \mathbb{E}_{x|a}[\varepsilon(x)].
\end{equation}

\paragraph{Proof.}
Taking expectation over $x|a$ in the allocation condition yields
\begin{equation}
R_a(\epsilon) = \mathbb{E}_{x|a} \left[\sum_j g_{\zeta}^{(j)}(x)L_j(x)\right]
\le \mathbb{E}_{x|a} \left[\min_j L_j(x)\right] + \mathbb{E}_{x|a}[\varepsilon(x)].
\end{equation}
For any $a,a'$, applying the triangle inequality gives
\begin{align}
|R_a(\epsilon)-R_{a'}(\epsilon)|
&\le \left| \mathbb{E}_{x|a}[\min_j L_j(x)] - \mathbb{E}_{x|a'}[\min_j L_j(x)] \right| \\
&\quad + \mathbb{E}_{x|a}[\varepsilon(x)] + \mathbb{E}_{x|a'}[\varepsilon(x)].
\end{align}
Taking the maximum over $(a,a')$ and bounding by $\bar{\varepsilon}$ yields the result.
\hfill $\square$

\end{document}

%% file: legend.tex
\begin{tikzpicture} 
    \begin{axis}[
        height=5em,
        width=0.9\linewidth,
        hide axis,
        xmin=10,
        xmax=50,
        ymin=0,
        ymax=0.4,
        legend style={draw=none, legend cell align=left, font={\scriptsize}, {/tikz/every even column/.append style={column sep=2em}}},
        legend columns=-1
    ]

    % \addlegendimage{mark=star, mark options={scale=2, draw=Brown, fill=Brown, style=solid}, draw=Brown, style={dotted, thin}};
    % \addlegendentry{DaF};

    % \addlegendimage{mark=pentagon*, mark options={scale=1, draw=Purple, fill=Purple, style=solid}, draw=Purple, style={dashdotdotted, thin}};
    % \addlegendentry{A-SM};
    
    \addlegendimage{mark=pentagon*, mark options={scale=2., draw=MidnightBlue, fill=MidnightBlue, style=solid}, draw=MidnightBlue, style={dashdotdotted, thin}};
    \addlegendentry{DCE};

    \addlegendimage{mark=triangle*, mark options={scale=2., draw=BurntOrange, fill=BurntOrange, style=solid}, draw=BurntOrange, style={dashdotted, thin}};
    \addlegendentry{Fair L2D};

    \addlegendimage{mark=diamond*, mark options={scale=1.5, draw=Brown, fill=Brown, style=solid}, draw=Brown, style={dotted, thin}};
    \addlegendentry{LECODU};

    \addlegendimage{mark=square*, mark options={scale=1.5, draw=BrickRed, fill=BrickRed, style=solid}, draw=BrickRed, style={dotted, thin}};
    \addlegendentry{FairDi};

    \addlegendimage{mark=*, mark options={scale=1.5, draw=PineGreen, fill=PineGreen, style=solid}, draw=PineGreen, style={solid, thin}};
    \addlegendentry{PecMan};
    \end{axis}
\end{tikzpicture}

%% file: ham_gender_auc.tex
\begin{tikzpicture}
    \begin{axis}[
        height = 0.2\linewidth,
        width = 0.2\linewidth,
        xlabel={\empty},
        xticklabel={\empty},
        ylabel={AUC},
        xlabel style={font=\footnotesize},
        ylabel style={font=\footnotesize, yshift=-0.5em},
        xticklabel style={font=\scriptsize},
        yticklabel style={font=\scriptsize},
        % xtick distance={0.25},
        % ytick distance={0.05},
        ymin=0.8,
        ymax=1.,
        legend image post style={scale=1},
        legend cell align={left},
        legend pos=south west,
        legend columns=1,
        legend style={draw=none, font={\footnotesize}, yshift=0em, xshift=0.em},
        scale only axis
    ]
        
        \addplot[mark=pentagon*, mark options={scale=1, draw=MidnightBlue, fill=MidnightBlue, style=solid}, draw=MidnightBlue, style={dashdotdotted, thin},
        error bars/.cd,
        y dir=both,
        y explicit,
        error bar style={solid, MidnightBlue, thick, line width=0.5pt}] coordinates {
            (0., 0.9747)
            (0.2063, 0.9757) +- (0, 0.0022)
            (0.3674, 0.9815) +- (0, 0.0031)
            (0.6009, 0.9311) +- (0, 0.0053)
            (0.7665, 0.8721) +- (0, 0.0066)
            (1., 0.7041) +- (0, 0.0088)
        };
        % \addlegendentry{DCE};

        \addplot[mark=triangle*, mark options={scale=1, draw=BurntOrange, fill=BurntOrange, style=solid}, draw=BurntOrange, style={dashdotted, thin},
        error bars/.cd,
        y dir=both,
        y explicit,
        error bar style={solid, BurntOrange, thick, line width=0.5pt}] coordinates {
            (0, 0.9747)
            (0.1461, 0.9748) +- (0, 0.0052)
            (0.4819, 0.9404) +- (0, 0.0031)
            (0.5898, 0.9204) +- (0, 0.0051)
            (0.8835, 0.9139) +- (0, 0.0024)
            (1., 0.9060) +- (0, 0.0044)
        };
	% \addlegendentry{FairL2D};

        \addplot[mark=diamond*, mark options={scale=1, draw=Brown, fill=Brown, style=solid}, draw=Brown, style={dashed, thin},
        error bars/.cd,
        y dir=both,
        y explicit,
        error bar style={solid, Brown, thick, line width=0.5pt}] coordinates {
            (0., 0.9755) +- (0, 0.0002)
            (0.2011, 0.9748)  +- (0, 0.0045)
            (0.418, 0.9544)  +- (0, 0.0031)
            (0.5898, 0.9324)  +- (0, 0.0047)
            (0.8077, 0.9259)  +- (0, 0.0013)
            (1., 0.9060)  +- (0, 0.0012)
        };
	% \addlegendentry{LECODU};

        \addplot[mark=square*, mark options={scale=1, draw=BrickRed, fill=BrickRed, style=solid}, draw=BrickRed, style={dotted, thin}] coordinates {
            (0.0, 0.9747)
            (1., 0.9325)
        };
	% \addlegendentry{FairDi};

        \addplot[mark=*, mark options={scale=1, draw=PineGreen, fill=PineGreen, style=solid}, draw=PineGreen, style={solid, thin},
        error bars/.cd,
        y dir=both,
        y explicit,
        error bar style={solid, PineGreen, thick, line width=0.5pt}] coordinates {
            (0, 0.9866) +- (0, 0.0016)
            (0.1358, 0.9861) +- (0, 0.0016)
            (0.4302, 0.9882) +- (0, 0.0032)
            (0.5998, 0.9772) +- (0, 0.0049)
            (0.8529, 0.9582) +- (0, 0.0011)
            (1., 0.9470) +- (0, 0.0021)
        };
	% \addlegendentry{PecMan};
    \end{axis}
\end{tikzpicture}

%% file: CMMD_auc.tex
\begin{tikzpicture}
    \begin{axis}[
        height = 0.2\linewidth,
        width = 0.2\linewidth,
        xlabel={\empty},
        xticklabel={\empty},
        xlabel style={font=\small},
        xticklabel style={font=\small},
        yticklabel style={font=\small},
        ymin=0.8,
        ymax=1.,
        % xtick distance={0.25},
        % ytick distance={0.05},
        % xmin=-0.05,
        % xmax=1.05,
        yticklabel={\empty},
        legend image post style={scale=1},
        legend cell align={left},
        legend pos=south west,
        legend columns=1,
        legend style={draw=none, font={\footnotesize}, yshift=0em, xshift=0.em},
        scale only axis
    ]

        \addplot[mark=pentagon*, mark options={scale=1, draw=MidnightBlue, fill=MidnightBlue, style=solid}, draw=MidnightBlue, style={dashdotdotted, thin},
        error bars/.cd,
        y dir=both,
        y explicit,
        error bar style={solid, MidnightBlue, thick, line width=0.5pt}] coordinates {
            (0., 0.9378)
            (0.1867, 0.9281)  +- (0, 0.0069)
            (0.4046, 0.8715)  +- (0, 0.0044)
            (0.5661, 0.8629)  +- (0, 0.0082)
            (0.8190, 0.8175)  +- (0, 0.0065)
            (1., 0.7175)  +- (0, 0.0077)
        };
        % \addlegendentry{DCE};

        \addplot[mark=triangle*, mark options={scale=1, draw=BurntOrange, fill=BurntOrange, style=solid}, draw=BurntOrange, style={dashdotted, thin},
        error bars/.cd,
        y dir=both,
        y explicit,
        error bar style={solid, BurntOrange, thick, line width=0.5pt}] coordinates {
            (0, 0.9378)
            (0.1643, 0.9371)  +- (0, 0.0025)
            (0.4114, 0.9297)  +- (0, 0.0032)
            (0.5778, 0.9197)  +- (0, 0.0029)
            (0.7723, 0.9072)  +- (0, 0.0028)
            (1., 0.8787) +- (0, 0.0012)
        };
	% \addlegendentry{FairL2D};

        \addplot[mark=diamond*, mark options={scale=1, draw=Brown, fill=Brown, style=solid}, draw=Brown, style={dashed, thin},
        error bars/.cd,
        y dir=both,
        y explicit,
        error bar style={solid, Brown, thick, line width=0.5pt}] coordinates {
            (0., 0.9390) +- (0, 0.0022)
            (0.1711, 0.9359)  +- (0, 0.0055)
            (0.4218, 0.9300)  +- (0, 0.0061)
            (0.5698, 0.9188)  +- (0, 0.0047)
            (0.7577, 0.9172)  +- (0, 0.0053)
            (1., 0.8787)  +- (0, 0.0012)
        };
	% \addlegendentry{LECODU};

        \addplot[mark=square*, mark options={scale=1, draw=BrickRed, fill=BrickRed, style=solid}, draw=BrickRed, style={dotted, thin}] coordinates {
            (0.0, 0.9378)
            (1., 0.8829)
        };
	% \addlegendentry{FairDi};

        \addplot[mark=*, mark options={scale=1, draw=PineGreen, fill=PineGreen, style=solid}, draw=PineGreen, style={solid, thin},
        error bars/.cd,
        y dir=both,
        y explicit,
        error bar style={solid, PineGreen, thick, line width=0.5pt}] coordinates {
            (0, 0.9468) +- (0, 0.0013)
            (0.1770, 0.9405) +- (0, 0.0017)
            (0.4036, 0.9363) +- (0, 0.0042)
            (0.6215, 0.9305) +- (0, 0.0019)
            (0.8959, 0.9093) +- (0, 0.0037)
            (1., 0.8898) +- (0, 0.0034)
        };
	% \addlegendentry{Pecman};

    \end{axis}
\end{tikzpicture}

%% file: CXP_auc.tex
\begin{tikzpicture}
    \begin{axis}[
        height = 0.2\linewidth,
        width = 0.2\linewidth,
        xlabel={\empty},
        xticklabel={\empty},
        xlabel style={font=\small},
        xticklabel style={font=\small},
        % yticklabel style={font=\small},
        % xtick distance={0.25},
        % ytick distance={0.05},
        % xmin=-0.05,
        % xmax=1.05,
        ymin=0.8,
        ymax=1.,
        yticklabel={\empty},
        legend image post style={scale=1},
        legend cell align={left},
        legend pos=south west,
        legend columns=1,
        legend style={draw=none, font={\footnotesize}, yshift=0em, xshift=0.em},
        scale only axis
    ]

        \addplot[mark=diamond*, mark options={scale=1, draw=Brown, fill=Brown, style=solid}, draw=Brown, style={dashed, thin},
        error bars/.cd,
        y dir=both,
        y explicit,
        error bar style={solid, Brown, thick, line width=0.5pt}] coordinates {
            (0., 0.9515)
            (0.2069, 0.9322)  +- (0, 0.0045)
            (0.318, 0.9354)  +- (0, 0.0032)
            (0.5598, 0.9347)  +- (0, 0.0017)
            (0.8077, 0.9015)  +- (0, 0.0033)
            (1., 0.8727)  +- (0, 0.0006)
        };
	% \addlegendentry{LECODU};

        \addplot[mark=pentagon*, mark options={scale=1, draw=MidnightBlue, fill=MidnightBlue, style=solid}, draw=MidnightBlue, style={dashdotdotted, thin},
        error bars/.cd,
        y dir=both,
        y explicit,
        error bar style={solid, MidnightBlue, thick, line width=0.5pt}] coordinates {
            (0, 0.9515)
            (0.1974, 0.9457)  +- (0, 0.0019)
            (0.3875, 0.9462)  +- (0, 0.0032)
            (0.6076, 0.9214)  +- (0, 0.0011)
            (0.7961, 0.8439)  +- (0, 0.0047)
            (1., 0.8012)  +- (0, 0.0056)
        };
        % \addlegendentry{DCE};

        \addplot[mark=triangle*, mark options={scale=1, draw=BurntOrange, fill=BurntOrange, style=solid}, draw=BurntOrange, style={dashdotted, thin},
        error bars/.cd,
        y dir=both,
        y explicit,
        error bar style={solid, BurntOrange, thick, line width=0.5pt}] coordinates {
            (0, 0.9515) 
            (0.1588, 0.9314)  +- (0, 0.0010)
            (0.4016, 0.9204)  +- (0, 0.0046)
            (0.5729, 0.9147)  +- (0, 0.0012)
            (0.7881, 0.9093)  +- (0, 0.0038)
            (1., 0.8727)  +- (0, 0.0043)
        };
	% \addlegendentry{FairL2D};

        \addplot[mark=square*, mark options={scale=1, draw=BrickRed, fill=BrickRed, style=solid}, draw=BrickRed, style={dotted, thin},
        error bars/.cd,
        y dir=both,
        y explicit,
        error bar style={solid, BrickRed, thick, line width=0.5pt}] coordinates {
            (0.0, 0.9515)
            (1., 0.8752)
        };
	% \addlegendentry{FairDi};

        \addplot[mark=*, mark options={scale=1, draw=PineGreen, fill=PineGreen, style=solid}, draw=PineGreen, style={solid, thin},
        error bars/.cd,
        y dir=both,
        y explicit,
        error bar style={solid, PineGreen, thick, line width=0.5pt}] coordinates {
            (0, 0.9585)  +- (0, 0.0016)
            (0.1926, 0.9527)  +- (0, 0.0026)
            (0.3845, 0.9512)  +- (0, 0.0037)
            (0.6013, 0.9421)  +- (0, 0.0011)
            (0.7769, 0.9213)  +- (0, 0.0045)
            (1., 0.8903)  +- (0, 0.0032)
        };
	% \addlegendentry{PecMan};
    \end{axis}
\end{tikzpicture}

%% file: MIMIC_auc.tex
\begin{tikzpicture}
    \begin{axis}[
        height = 0.2\linewidth,
        width = 0.2\linewidth,
        xlabel={\empty},
        xticklabel={\empty},
        xlabel style={font=\small},
        xticklabel style={font=\small},
        ymin=0.8,
        ymax=1.,
        % xtick distance={0.25},
        % ytick distance={0.05},
        % xmin=-0.05,
        % xmax=1.05,
        yticklabel={\empty},
        legend image post style={scale=1},
        legend cell align={left},
        legend pos=south west,
        legend columns=1,
        legend style={draw=none, font={\footnotesize}, yshift=0em, xshift=0.em},
        scale only axis
    ]

        \addplot[mark=pentagon*, mark options={scale=1, draw=MidnightBlue, fill=MidnightBlue, style=solid}, draw=MidnightBlue, style={dashdotdotted, thin},
        error bars/.cd,
        y dir=both,
        y explicit,
        error bar style={solid, Brown, thick, line width=0.5pt}] coordinates {
            (0., 0.9476)
            (0.19, 0.9481)  +- (0, 0.0012)
            (0.3977, 0.9273)  +- (0, 0.0038)
            (0.5963, 0.9140)  +- (0, 0.0052)
            (0.7553, 0.8802)  +- (0, 0.0037)
            (1., 0.8116)  +- (0, 0.0079)
        };
        % \addlegendentry{DCE};

        \addplot[mark=triangle*, mark options={scale=1, draw=BurntOrange, fill=BurntOrange, style=solid}, draw=BurntOrange, style={dashdotted, thin},
        error bars/.cd,
        y dir=both,
        y explicit,
        error bar style={solid, Brown, thick, line width=0.5pt}] coordinates {
            (0, 0.9476)
            (0.1612, 0.9411)  +- (0, 0.0019)
            (0.4126, 0.9153)  +- (0, 0.0023)
            (0.6001, 0.9131)  +- (0, 0.0042)
            (0.7808, 0.8785)  +- (0, 0.0027)
            (1., 0.8626)  +- (0, 0.0006)
        };
	% \addlegendentry{FairL2D};

    \addplot[mark=diamond*, mark options={scale=1, draw=Brown, fill=Brown, style=solid}, draw=Brown, style={dashed, thin},
        error bars/.cd,
        y dir=both,
        y explicit,
        error bar style={solid, Brown, thick, line width=0.5pt}] coordinates {
            (0., 0.9463)  +- (0, 0.0027)
            (0.2069, 0.9429)  +- (0, 0.0045)
            (0.418, 0.9118)  +- (0, 0.0032)
            (0.5898, 0.9151)  +- (0, 0.0017)
            (0.77, 0.8882)  +- (0, 0.0033)
            (1., 0.8626)  +- (0, 0.0006)
        };
	% \addlegendentry{LECODU};

        \addplot[mark=square*, mark options={scale=1, draw=BrickRed, fill=BrickRed, style=solid}, draw=BrickRed, style={dotted, thin}]  coordinates {
            (0.0, 0.9476)
            (1., 0.8786)
        };
	% \addlegendentry{FairDi};

        \addplot[mark=*, mark options={scale=1, draw=PineGreen, fill=PineGreen, style=solid}, draw=PineGreen, style={solid, thin},
        error bars/.cd,
        y dir=both,
        y explicit,
        error bar style={solid, Brown, thick, line width=0.5pt}] coordinates {
            (0, 0.9616)  +- (0, 0.0008)
            (0.1721, 0.9565)  +- (0, 0.0009)
            (0.39, 0.9390)  +- (0, 0.0025)
            (0.5977, 0.9262)  +- (0, 0.0041)
            (0.7931, 0.9066)  +- (0, 0.0006)
            (1., 0.8892)  +- (0, 0.0012)
        };
	% \addlegendentry{Pecman};

    \end{axis}
\end{tikzpicture}

%% file: ham_gender_es_auc.tex
\begin{tikzpicture}
    \begin{axis}[
        height = 0.2\linewidth,
        width = 0.2\linewidth,
        xlabel={Coverage},
        ylabel={ES-AUC},
        xlabel style={font=\footnotesize},
        ylabel style={font=\footnotesize, yshift=-0.5em},
        xticklabel style={font=\scriptsize},
        yticklabel style={font=\scriptsize},
        ytick scale label code/.code={}, 
        scaled y ticks=false, 
        % xtick distance={0.25},
        % xmin=-0.05,
        % xmax=1.05,
        ymin=0.8,
        ymax=1.,
        legend image post style={scale=1},
        legend cell align={left},
        legend pos=north west,
        legend columns=1,
        legend style={draw=none, font={\footnotesize}, yshift=0em, xshift=0.em},
        scale only axis
    ]

        \addplot[mark=pentagon*, mark options={scale=1, draw=MidnightBlue, fill=MidnightBlue, style=solid}, draw=MidnightBlue, style={dashdotdotted, thin},
        error bars/.cd,
        y dir=both,
        y explicit,
        error bar style={solid, MidnightBlue, thick, line width=0.5pt}] coordinates {
            (0., 0.9716)
            (0.2063, 0.9736) +- (0, 0.0039)
            (0.3674, 0.9749) +- (0, 0.0028)
            (0.6009, 0.9084) +- (0, 0.0047)
            (0.7665, 0.8572) +- (0, 0.0032)
            (1., 0.7041) +- (0, 0.0065)
        };

        \addplot[mark=triangle*, mark options={scale=1, draw=BurntOrange, fill=BurntOrange, style=solid}, draw=BurntOrange, style={dashdotted, thin},
        error bars/.cd,
        y dir=both,
        y explicit,
        error bar style={solid, BurntOrange, thick, line width=0.5pt}] coordinates {
            (0, 0.9716)
            (0.1461, 0.9699) +- (0, 0.0045)
            (0.4819, 0.9365) +- (0, 0.0034)
            (0.5898, 0.9178) +- (0, 0.0041)
            (0.8835, 0.9110) +- (0, 0.0015)
            (1., 0.9020) +- (0, 0.0036)
        };

        \addplot[mark=diamond*, mark options={scale=1, draw=Brown, fill=Brown, style=solid}, draw=Brown, style={dashed, thin},
        error bars/.cd,
        y dir=both,
        y explicit,
        error bar style={solid, Brown, thick, line width=0.5pt}] coordinates {
            (0., 0.9745) +- (0, 0.0012)
            (0.2011, 0.9738)  +- (0, 0.0045)
            (0.418, 0.9532)  +- (0, 0.0031)
            (0.5898, 0.9304)  +- (0, 0.0047)
            (0.8077, 0.9239)  +- (0, 0.0013)
            (1., 0.9030)  +- (0, 0.0012)
        };
	% \addlegendentry{LECODU};

        \addplot[mark=square*, mark options={scale=1, draw=BrickRed, fill=BrickRed, style=solid}, draw=BrickRed, style={dotted, thin}] coordinates {
            (0.0, 0.9716)
            (1., 0.9314)
        };

        \addplot[mark=*, mark options={scale=1, draw=PineGreen, fill=PineGreen, style=solid}, draw=PineGreen, style={solid, thin},
        error bars/.cd,
        y dir=both,
        y explicit,
        error bar style={solid, PineGreen, thick, line width=0.5pt}] coordinates {
            (0, 0.9816) +- (0, 0.0016)
            (0.1358, 0.9820) +- (0, 0.0016)
            (0.4302, 0.9830) +- (0, 0.0012)
            (0.5999, 0.9732) +- (0, 0.0023)
            (0.8529, 0.9544) +- (0, 0.0011)
            (1., 0.9444) +- (0, 0.0021)
        };

    \end{axis}
\end{tikzpicture}

%% file: CMMD_esauc.tex
\begin{tikzpicture}
    \begin{axis}[
        height = 0.2\linewidth,
        width = 0.2\linewidth,
        xlabel={Coverage},
        xlabel style={font=\footnotesize},
        xticklabel style={font=\scriptsize},
        yticklabel style={font=\scriptsize},
        ytick scale label code/.code={}, 
        scaled y ticks=false, 
        % xtick distance={0.25},
        % xmin=-0.05,
        % xmax=1.05,
        ymin=0.8,
        ymax=1.,
        yticklabel={\empty},
        legend image post style={scale=1},
        legend cell align={left},
        legend pos=north west,
        legend columns=1,
        legend style={draw=none, font={\footnotesize}, yshift=0em, xshift=0.em},
        scale only axis
    ]

        \addplot[mark=pentagon*, mark options={scale=1, draw=MidnightBlue, fill=MidnightBlue, style=solid}, draw=MidnightBlue, style={dashdotdotted, thin},
        error bars/.cd,
        y dir=both,
        y explicit,
        error bar style={solid, MidnightBlue, thick, line width=0.5pt}] coordinates {
            (0., 0.9079)
            (0.1867, 0.89) +- (0, 0.0029)
            (0.4046, 0.8389) +- (0, 0.0056)
            (0.5661, 0.8246) +- (0, 0.0073)
            (0.8190, 0.7779) +- (0, 0.0032)
            (1., 0.6779) +- (0, 0.0066)
        };
        % \addlegendentry{DCE};

        \addplot[mark=triangle*, mark options={scale=1, draw=BurntOrange, fill=BurntOrange, style=solid}, draw=BurntOrange, style={dashdotted, thin},
        error bars/.cd,
        y dir=both,
        y explicit,
        error bar style={solid, BurntOrange, thick, line width=0.5pt}] coordinates {
            (0, 0.9079)
            (0.1643, 0.9058) +- (0, 0.0079)
            (0.4114, 0.8967) +- (0, 0.0052)
            (0.5778, 0.8895) +- (0, 0.0044)
            (0.7723, 0.8727) +- (0, 0.0041)
            (1., 0.8317) +- (0, 0.0033)
        };

        \addplot[mark=diamond*, mark options={scale=1, draw=Brown, fill=Brown, style=solid}, draw=Brown, style={dashed, thin},
        error bars/.cd,
        y dir=both,
        y explicit,
        error bar style={solid, Brown, thick, line width=0.5pt}] coordinates {
            (0., 0.9099) +- (0, 0.0022)
            (0.1711, 0.9068)  +- (0, 0.0055)
            (0.4218, 0.8972)  +- (0, 0.0061)
            (0.5698, 0.8905)  +- (0, 0.0047)
            (0.7577, 0.8737)  +- (0, 0.0053)
            (1., 0.8327)  +- (0, 0.0012)
        };
	% \addlegendentry{LECODU};

        \addplot[mark=square*, mark options={scale=1, draw=BrickRed, fill=BrickRed, style=solid}, draw=BrickRed, style={dotted, thin}] coordinates {
            (0.0, 0.9079)
            (1., 0.8436)
        };

        \addplot[mark=*, mark options={scale=1, draw=PineGreen, fill=PineGreen, style=solid}, draw=PineGreen, style={solid, thin},
        error bars/.cd,
        y dir=both,
        y explicit,
        error bar style={solid, PineGreen, thick, line width=0.5pt}] coordinates {
            (0, 0.9123) +- (0, 0.0020)
            (0.1770, 0.9111) +- (0, 0.0040)
            (0.4036, 0.9057) +- (0, 0.0039)
            (0.6215, 0.9040) +- (0, 0.0033)
            (0.8959, 0.8681) +- (0, 0.0048)
            (1., 0.8515) +- (0, 0.0019)
        };

    \end{axis}
\end{tikzpicture}

%% file: CXP_esauc.tex
\begin{tikzpicture}
    \begin{axis}[
        height = 0.2\linewidth,
        width = 0.2\linewidth,
        xlabel={Coverage},
        xlabel style={font=\footnotesize},
        xticklabel style={font=\scriptsize},
        ytick scale label code/.code={}, 
        scaled y ticks=false, 
        % xtick distance={0.25},
        % xmin=-0.05,
        % xmax=1.05,
        ymin=0.8,
        ymax=1.,
        yticklabel={\empty},
        legend image post style={scale=1},
        legend cell align={left},
        legend pos=north west,
        legend columns=1,
        legend style={draw=none, font={\footnotesize}, yshift=0em, xshift=0.em},
        scale only axis
    ]

        \addplot[mark=diamond*, mark options={scale=1, draw=Brown, fill=Brown, style=solid}, draw=Brown, style={dashed, thin},
        error bars/.cd,
        y dir=both,
        y explicit,
        error bar style={solid, Brown, thick, line width=0.5pt}] coordinates {
            (0., 0.9505)
            (0.2069, 0.9318)  +- (0, 0.0045)
            (0.318, 0.9343) +- (0, 0.0032)
            (0.5598, 0.9338)  +- (0, 0.0017)
            (0.8077, 0.8988)  +- (0, 0.0035)
            (1., 0.8676)  +- (0, 0.0041)
        };

        \addplot[mark=pentagon*, mark options={scale=1, draw=MidnightBlue, fill=MidnightBlue, style=solid}, draw=MidnightBlue, style={dashdotdotted, thin},
        error bars/.cd,
        y dir=both,
        y explicit,
        error bar style={solid, MidnightBlue, thick, line width=0.5pt}] coordinates {
            (0, 0.9505)
            (0.1974, 0.9454)  +- (0, 0.0019)
            (0.3875, 0.9437)  +- (0, 0.0032)
            (0.6076, 0.9165)  +- (0, 0.0011)
            (0.7961, 0.8354)  +- (0, 0.0047)
            (1., 0.7966)  +- (0, 0.0056)
        };

        \addplot[mark=triangle*, mark options={scale=1, draw=BurntOrange, fill=BurntOrange, style=solid}, draw=BurntOrange, style={dashdotted, thin},
        error bars/.cd,
        y dir=both,
        y explicit,
        error bar style={solid, BurntOrange, thick, line width=0.5pt}] coordinates {
            (0, 0.9505)
            (0.2009, 0.9310)  +- (0, 0.0010)
            (0.4016, 0.9196)  +- (0, 0.0042)
            (0.5729, 0.9094)  +- (0, 0.0016)
            (0.7881, 0.9064)  +- (0, 0.0038)
            (1., 0.8676)  +- (0, 0.0043)
        };

        \addplot[mark=square*, mark options={scale=1, draw=BrickRed, fill=BrickRed, style=solid}, draw=BrickRed, style={dotted, thin},
        error bars/.cd,
        y dir=both,
        y explicit,
        error bar style={solid, BrickRed, thick, line width=0.5pt}] coordinates {
            (0.0, 0.9505)
            (1., 0.8744)
        };

        \addplot[mark=*, mark options={scale=1, draw=PineGreen, fill=PineGreen, style=solid}, draw=PineGreen, style={solid, thin},
        error bars/.cd,
        y dir=both,
        y explicit,
        error bar style={solid, PineGreen, thick, line width=0.5pt}] coordinates {
            (0, 0.9565)  +- (0, 0.0026)
            (0.1926, 0.9511)  +- (0, 0.0026)
            (0.3845, 0.9495)  +- (0, 0.0037)
            (0.6013, 0.9381)  +- (0, 0.0011)
            (0.7769, 0.9212)  +- (0, 0.0045)
            (1., 0.8845)  +- (0, 0.0032)
        };

    \end{axis}
\end{tikzpicture}

%% file: MIMIC_esauc.tex
\begin{tikzpicture}
    \begin{axis}[
        height = 0.2\linewidth,
        width = 0.2\linewidth,
        xlabel={Coverage},
        xlabel style={font=\footnotesize},
        xticklabel style={font=\scriptsize},
        yticklabel style={font=\scriptsize},
        ytick scale label code/.code={}, 
        scaled y ticks=false, 
        % xtick distance={0.25},
        % xmin=-0.05,
        % xmax=1.05,
        ymin=0.8,
        ymax=1.,
        yticklabel={\empty},
        legend image post style={scale=1},
        legend cell align={left},
        legend pos=north west,
        legend columns=1,
        legend style={draw=none, font={\footnotesize}, yshift=0em, xshift=0.em},
        scale only axis
    ]

        \addplot[mark=pentagon*, mark options={scale=1, draw=MidnightBlue, fill=MidnightBlue, style=solid}, draw=MidnightBlue, style={dashdotdotted, thin}] coordinates {
            (0., 0.9465) 
            (0.1867, 0.9470) +- (0, 0.0046)
            (0.3977, 0.9254) +- (0, 0.0029)
            (0.5963, 0.9125) +- (0, 0.0021)
            (0.7617, 0.8756) +- (0, 0.0057)
            (1., 0.8102) +- (0, 0.0028)
        };
        % \addlegendentry{DCE};

        \addplot[mark=triangle*, mark options={scale=1, draw=BurntOrange, fill=BurntOrange, style=solid}, draw=BurntOrange, style={dashdotted, thin}] coordinates {
            (0, 0.9465)
            (0.19, 0.9470) +- (0, 0.0019)
            (0.4126, 0.9085) +- (0, 0.0026)
            (0.6001, 0.9113) +- (0, 0.0002)
            (0.7808, 0.8750) +- (0, 0.0044)
            (1., 0.8589) +- (0, 0.0006)
        };

        \addplot[mark=diamond*, mark options={scale=1, draw=Brown, fill=Brown, style=solid}, draw=Brown, style={dashed, thin},
        error bars/.cd,
        y dir=both,
        y explicit,
        error bar style={solid, Brown, thick, line width=0.5pt}] coordinates {
            (0., 0.9443)  +- (0, 0.0027)
            (0.2069, 0.9417)  +- (0, 0.0045)
            (0.418, 0.9112)  +- (0, 0.0032)
            (0.5898, 0.9111)  +- (0, 0.0017)
            (0.77, 0.8852)  +- (0, 0.0033)
            (1., 0.8589)  +- (0, 0.0006)
        };
	% \addlegendentry{LECODU};

        \addplot[mark=square*, mark options={scale=1, draw=BrickRed, fill=BrickRed, style=solid}, draw=BrickRed, style={dotted, thin}]  coordinates {
            (0.0, 0.9465)
            (1., 0.8744)
        };

        \addplot[mark=*, mark options={scale=1, draw=PineGreen, fill=PineGreen, style=solid}, draw=PineGreen, style={solid, thin}] coordinates {
            (0, 0.9565) +- (0, 0.0011)
            (0.1721, 0.9522) +- (0, 0.0012)
            (0.39, 0.9313) +- (0, 0.0029)
            (0.5977, 0.9172) +- (0, 0.0017)
            (0.7931, 0.8961) +- (0, 0.0032)
            (1., 0.8789) +- (0, 0.0006)
        };

    \end{axis}
\end{tikzpicture}

%% file: deferral.tex
\begin{tikzpicture}
    \begin{axis}[
        height = 0.25\linewidth,
        width = 0.25\linewidth,
        ybar=1pt,
        bar width=6pt,
        axis x line=bottom,
        axis y line=left,
        ymax=95,
        ymin=12,
        xlabel={ \(1 -\varepsilon\)},
        ylabel={Deferral rate (\%)},
        xlabel style={font=\footnotesize, yshift=1ex},
        ylabel style={font=\footnotesize, yshift=-1ex},
        xticklabel style={font=\scriptsize, align=center},
        yticklabel style={font=\scriptsize},
        % symbolic x coords={Expert-1 91\%, Expert-2 87\%, AI classifier 82\%},
        % xtick distance={1},
        % xticklabels={Expert 1, Expert 1\\91\%, Expert 2\\87\%, AI classifier\\82\%},
        legend image post style={scale=0.8},
        legend columns=2,
        legend cell align={left},
        legend pos=north west,
        legend style={draw=none, font={\scriptsize}, xshift=-0.275em, yshift=1ex, text opacity=1, fill opacity=0.75},
        enlarge x limits=0.25,
        nodes near coords,
        nodes near coords align={vertical},
        nodes near coords style={font=\scriptsize, rotate=90, anchor=west, /pgf/number format/.cd, fixed zerofill, precision=2},
        scale only axis
    ]
        % cohort 1
        \addplot[mark=none, draw=MidnightBlue, fill=MidnightBlue] coordinates {
            (0.2, 40.28)
            (0.4, 33.36)
            (0.6, 26.61)
        };
        \addlegendentry{AI-male};

        % cohort 2
        \addplot[mark=none, draw=BurntOrange, fill=BurntOrange] coordinates {
            (0.2, 45.11)
            (0.4, 24.80)
            (0.6, 15.31)
        };
        \addlegendentry{AI-female};

        % human
        \addplot[mark=none, draw=PineGreen, fill=PineGreen] coordinates {
            (0.2, 14.61)
            (0.4, 41.84)
            (0.6, 58.08)
        };
        \addlegendentry{Clinician};
    \end{axis}
\end{tikzpicture}

%% file: confusion_matrix.tex
\begin{tikzpicture}
    \begin{axis}[
        height = 0.225 \linewidth,
        width = 0.25 \linewidth,
        xtick distance={1},
        xticklabels={dummy, {AI-male}, {\vphantom{AI}\\AI-female}, Clinician},
        xtick align={center},
        xlabel={\empty},
        % x tick label style={rotate=-30, anchor=west, yshift=-1ex},
        ytick distance={1},
        yticklabels={dummy, Male, Female},
        ytick align={center},
        y tick label style={rotate=90, anchor=south},
        xlabel style={font=\footnotesize},
        ylabel style={font=\footnotesize},
        xticklabel style={font=\scriptsize, align=center},
        yticklabel style={font=\scriptsize},
        % colorbar,
        colormap name={viridis},
        scale only axis,
        nodes near coords align={vertical},
        nodes near coords style={font=\scriptsize, yshift=-1.5ex, color=White},
        enlargelimits=false
    ]
        \addplot [
            matrix plot,
            nodes near coords,
            point meta min=5,
            point meta max=32,
            point meta=explicit
        ] coordinates {
            (0, 0) [19.18]  % [382]
            (1, 0) [6.88]  % [137]
            (2, 0) [23.39]  % [466]

            (0, 1) [7.68]  % [153]
            (1, 1) [15.56]  % [310]
            (2, 1) [27.31]  % [544]
        };
    \end{axis}
\end{tikzpicture}

%% file: auc_each_component.tex
\begin{tikzpicture}
    \begin{axis}[
        height = 0.25\linewidth,
        width = 0.275\linewidth,
        ybar=1pt,
        bar width=6pt,
        axis x line={bottom},
        axis y line={left},
        xlabel={\empty},
        ylabel={\empty},
        xlabel style={font=\footnotesize},
        ylabel style={font=\footnotesize},
        xticklabel style={font=\scriptsize, align=center},
        yticklabel style={font=\scriptsize},
        xtick distance={1},
        xticklabels={dummy, AI-male, \vphantom{AI}\\AI-female, Clinician},
        ymax={99},
        ymin={92},
        legend image post style={scale=0.75},
        legend columns=1,
        legend cell align={left},
        legend pos=north west,
        legend style={draw=none, font={\scriptsize}, yshift=0.8ex, text opacity=1, fill opacity=0.75},
        enlarge x limits=0.3,
        nodes near coords,
        nodes near coords align={vertical},
        nodes near coords style={font=\scriptsize, rotate=90, anchor=west, /pgf/number format/.cd, fixed zerofill, precision=2},
        scale only axis
    ]
        % AUC on Male
        \addplot[mark=none, draw=Gray, fill=Gray] coordinates {
            (0, 94.75)
            (1, 93.00)
            (2, 97.77)
            % (3, 93.26)
        };
        \addlegendentry{Male};

        % AUC on Female
        \addplot[mark=none, draw=Periwinkle, fill=Periwinkle] coordinates {
            (0, 93.45)
            (1, 94.85)
            (2, 97.04)
            % (3, 92.75)
        };
        \addlegendentry{Female};

        % AUC overall
        \addplot[mark=none, draw=TealBlue, fill=TealBlue] coordinates {
            (0, 93.87)
            (1, 93.61)
            (2, 97.47)
            % (3, 94.44)
        };
        \addlegendentry{Overall};
    \end{axis}
\end{tikzpicture}

%% file: ablation_coverage_legend.tex
\begin{tikzpicture} 
    \begin{axis}[
        height=5em,
        width=0.3\linewidth,
        hide axis,
        xmin=10,
        xmax=50,
        ymin=0,
        ymax=0.4,
        legend style={draw=none, legend cell align=left, font={\footnotesize}, {/tikz/every even column/.append style={column sep=1em}}},
        legend columns=-1
    ]
    
    \addlegendimage{mark=square, mark options={scale=1, draw=BurntOrange, fill=BurntOrange, style=solid}, draw=BurntOrange, style={dashdotted, thin}};
    \addlegendentry{L2D-FairDi};

    \addlegendimage{mark=*, mark options={scale=1, draw=PineGreen, fill=PineGreen, style=solid}, draw=PineGreen, style={solid, thin}};
    \addlegendentry{PecMan};
    \end{axis}
\end{tikzpicture}

%% file: ham_male.tex
\begin{tikzpicture}
    \begin{axis}[
        height = 0.22\linewidth,
        width = 0.22\linewidth,
        xlabel={Coverage},
        xlabel style={font=\footnotesize},
        ylabel style={font=\footnotesize, yshift=-0.5em},
        xticklabel style={font=\scriptsize},
        yticklabel style={font=\scriptsize},
        % xtick distance={0.2},
        % ytick distance={0.02},
        % xmin=-0.05,
        % xmax=1.05,
        % yticklabel={\empty},
        ymin=0.935,
        ymax=1.00,
        legend image post style={scale=1},
        legend cell align={left},
        legend pos=south west,
        legend columns=1,
        legend style={draw=none, font={\footnotesize}, yshift=0em, xshift=0.em},
        scale only axis
    ]
        \addplot[mark=square, mark options={scale=1, draw=BurntOrange, fill=BurntOrange, style=solid}, draw=BurntOrange, style={dashdotted, thin},
            error bars/.cd,
            y dir=both,
            y explicit,
            error bar style={solid, BurntOrange, thick, line width=0.5pt}] coordinates {
            (0.0, 0.9777)
            (0.1596, 0.9747) +- (0, 0.0016)
            (0.3599, 0.9655) +- (0, 0.0032)
            (0.5707, 0.9636) +- (0, 0.0041)
            (0.8473, 0.9551) +- (0, 0.0029)
            (1., 0.9406)
        };

         \addplot[mark=*, mark options={scale=1, draw=PineGreen, fill=PineGreen, style=solid}, draw=PineGreen, style={solid, thin},
            error bars/.cd,
            y dir=both,
            y explicit,
            error bar style={solid, PineGreen, thick, line width=0.5pt}] coordinates {
                (0, 0.9845) +- (0, 0.0016)
                (0.1358, 0.9811) +- (0, 0.0016)
                (0.4302, 0.9852) +- (0, 0.0032)
                (0.5998, 0.9712) +- (0, 0.0049)
                (0.8529, 0.9582) +- (0, 0.0011)
                (1., 0.9470) +- (0, 0.0021)
            };
        % \addlegendentry{PecMan};
        \end{axis}
\end{tikzpicture}

%% file: ham_female.tex
\begin{tikzpicture}
    \begin{axis}[
        height = 0.22\linewidth,
        width = 0.22\linewidth,
        xlabel={Coverage},
        xlabel style={font=\footnotesize},
        ylabel style={font=\footnotesize, yshift=-0.5em},
        xticklabel style={font=\scriptsize},
        yticklabel style={font=\scriptsize},
        % xtick distance={0.2},
        % ytick distance={0.02},
        % xmin=-0.05,
        % xmax=1.05,
        yticklabel={\empty},
        ymin=0.935,
        ymax=1.00,
        legend image post style={scale=1},
        legend cell align={left},
        legend pos=south west,
        legend columns=1,
        legend style={draw=none, font={\footnotesize}, yshift=0em, xshift=0.em},
        scale only axis
    ]
        \addplot[mark=square, mark options={scale=1, draw=BurntOrange, fill=BurntOrange, style=solid}, draw=BurntOrange, style={dashdotted, thin},
        error bars/.cd,
        y dir=both,
        y explicit,
        error bar style={solid, BurntOrange, thick, line width=0.5pt}] coordinates {
            (0.0, 0.9704)
            (0.1596, 0.9734) +- (0, 0.0016)
            (0.3599, 0.9691) +- (0, 0.0041)
            (0.5707, 0.9601) +- (0, 0.0032)
            (0.8473, 0.9511) +- (0, 0.0029)
            (1., 0.9406)
        };
	% \addlegendentry{FairDi+L2D};

        \addplot[mark=*, mark options={scale=1, draw=PineGreen, fill=PineGreen, style=solid}, draw=PineGreen, style={solid, thin},
        error bars/.cd,
        y dir=both,
        y explicit,
        error bar style={solid, PineGreen, thick, line width=0.5pt}] coordinates {
            (0, 0.9888) +- (0, 0.0016)
            (0.1358, 0.9891) +- (0, 0.0016)
            (0.4302, 0.9811) +- (0, 0.0032)
            (0.5998, 0.9800) +- (0, 0.0049)
            (0.8529, 0.9562) +- (0, 0.0011)
            (1., 0.9468) +- (0, 0.0021)
        };
	% \addlegendentry{Pecman};

    \end{axis}
\end{tikzpicture}

%% file: ablation_ham_gender_auc.tex
\begin{tikzpicture}
    \begin{axis}[
        height = 0.22\linewidth,
        width = 0.22\linewidth,
        xlabel={Coverage},
        xlabel style={font=\footnotesize},
        ylabel style={font=\footnotesize, yshift=-0.5em},
        xticklabel style={font=\scriptsize},
        yticklabel style={font=\scriptsize},
        % xtick distance={0.2},
        % ytick distance={0.02},
        % xmin=-0.05,
        % xmax=1.05,
        yticklabel={\empty},
        ymin=0.935,
        ymax=1.00,
        legend image post style={scale=1},
        legend cell align={left},
        legend pos=south west,
        legend columns=1,
        legend style={draw=none, font={\footnotesize}, yshift=0em, xshift=0.em},
        scale only axis
    ]
    
 %        \addplot[mark=star, mark options={scale=1, draw=Brown, fill=Brown, style=solid}, draw=Brown, style={dotted, thin}] coordinates {
 %            (0., 0.872)
 %            (0.2244, 0.8907)
 %            (0.4207, 0.8992)
 %            (0.5647, 0.8927)
 %            (0.8022, 0.8447)
 %            (1., 0.7041)
 %        };
 %        \addlegendentry{LCE};
        
 %        \addplot[mark=diamond*, mark options={scale=1, draw=BurntOrange, fill=BurntOrange, style=solid}, draw=BrickRed, style={dashed, thin}] coordinates {
 %            (0., 0.872)
 %            (0.194, 0.891)
 %            (0.3093, 0.8977)
 %            (0.4561, 0.9029)
 %            (0.58, 0.8985)
 %            (0.6997, 0.8795)
 %            (1., 0.7041)
 %        };
	% \addlegendentry{OvA};

 %        \addplot[mark=pentagon*, mark options={scale=1, draw=Purple, fill=Purple, style=solid}, draw=Purple, style={dashdotdotted, thin}] coordinates {
 %            (0., 0.9747)
 %            (0.2063, 0.9757)
 %            (0.3674, 0.9815)
 %            (0.6009, 0.9311)
 %            (0.7665, 0.8721)
 %            (1., 0.7041)
 %        };
 %        \addlegendentry{DCE};

 %        \addplot[mark=triangle*, mark options={scale=1, draw=MidnightBlue, fill=MidnightBlue, style=solid}, draw=MidnightBlue, style={dashdotted, thin}] coordinates {
 %            (0, 0.9747)
 %            (0.1461, 0.9748)
 %            (0.4819, 0.9404)
 %            (0.5898, 0.9204)
 %            (0.8835, 0.9139)
 %            (1., 0.9060)
 %        };
	% \addlegendentry{ERM+L2D};

        \addplot[mark=square, mark options={scale=1, draw=BurntOrange, fill=BurntOrange, style=solid}, draw=BurntOrange, style={dashdotted, thin},
        error bars/.cd,
        y dir=both,
        y explicit,
        error bar style={solid, BurntOrange, thick, line width=0.5pt}] coordinates {
            (0, 0.9747)
            (0.1596, 0.9795) +- (0, 0.0016)
            (0.3599, 0.9691) +- (0, 0.0029)
            (0.5707, 0.9632) +- (0, 0.0041)
            (0.8473, 0.9537) +- (0, 0.0032)
            (1., 0.9437)
        };
	% \addlegendentry{FairDi+L2D};

        \addplot[mark=*, mark options={scale=1, draw=PineGreen, fill=PineGreen, style=solid}, draw=PineGreen, style={solid, thin},
        error bars/.cd,
        y dir=both,
        y explicit,
        error bar style={solid, PineGreen, thick, line width=0.5pt}] coordinates {
            (0, 0.9866) +- (0, 0.0016)
            (0.1358, 0.9861) +- (0, 0.0016)
            (0.4302, 0.9882) +- (0, 0.0032)
            (0.5998, 0.9772) +- (0, 0.0049)
            (0.8529, 0.9582) +- (0, 0.0011)
            (1., 0.9470) +- (0, 0.0021)
        };
	% \addlegendentry{PecMan};

        % \addplot[mark=square*, mark options={scale=1, draw=BrickRed, fill=BrickRed, style=solid}, draw=BrickRed, style={solid, thin}] coordinates {
        %     (0., 0.872)
        %     (0.2509, 0.8981)
        %     (0.3825, 0.9034)
        %     (0.5847, 0.8914)
        %     (0.7593, 0.8586)
        %     (1., 0.7508)
        % };
        % \addlegendentry{Fair L2DC};
    \end{axis}
\end{tikzpicture}

%% file: ablation_ham_gender_esauc.tex
\begin{tikzpicture}
    \begin{axis}[
        height = 0.22\linewidth,
        width = 0.22\linewidth,
        xlabel={Coverage},
        xlabel style={font=\footnotesize},
        ylabel style={font=\footnotesize, yshift=-0.5em},
        xticklabel style={font=\scriptsize},
        yticklabel style={font=\scriptsize},
        % xtick distance={0.2},
        % ytick distance={0.02},
        % xmin=-0.05,
        % xmax=1.05,
        yticklabel={\empty},
        ymin=0.935,
        ymax=1.00,
        legend image post style={scale=1},
        legend cell align={left},
        legend pos=south west,
        legend columns=1,
        legend style={draw=none, font={\footnotesize}, yshift=0em, xshift=0.em},
        scale only axis
    ]
        \addplot[mark=square, mark options={scale=1, draw=BurntOrange, fill=BurntOrange, style=solid}, draw=BurntOrange, style={dashdotted, thin},
        error bars/.cd,
        y dir=both,
        y explicit,
        error bar style={solid, BurntOrange, thick, line width=0.5pt}] coordinates {
            (0, 0.9716)
            (0.1596, 0.9742) +- (0, 0.0016)
            (0.3599, 0.9673) +- (0, 0.0029)
            (0.5707, 0.9615) +- (0, 0.0041)
            (0.8473, 0.9517) +- (0, 0.0032)
            (1., 0.9407)
        };
	% \addlegendentry{FairDi+L2D};

        \addplot[mark=*, mark options={scale=1, draw=PineGreen, fill=PineGreen, style=solid}, draw=PineGreen, style={solid, thin},
        error bars/.cd,
        y dir=both,
        y explicit,
        error bar style={solid, PineGreen, thick, line width=0.5pt}] coordinates {
            (0, 0.9816) +- (0, 0.0016)
            (0.1358, 0.9820) +- (0, 0.0016)
            (0.4302, 0.9830) +- (0, 0.0012)
            (0.5999, 0.9732) +- (0, 0.0023)
            (0.8529, 0.9544) +- (0, 0.0011)
            (1., 0.9444) +- (0, 0.0021)
        };
	% \addlegendentry{Pecman};

    \end{axis}
\end{tikzpicture}